\documentclass[sn-apa]{sn-jnl}% APA Reference Style
\jyear{2023}%

\theoremstyle{thmstyleone}%
\newtheorem{theorem}{Theorem}%  meant for continuous numbers
\newtheorem{proposition}[theorem]{Proposition}% 

\theoremstyle{thmstylethree}%
\usepackage{mathtools} % amsmath with fixes and additions
\usepackage{amssymb}
\usepackage{amsthm}
\DeclareMathOperator*{\argmin}{arg\,min}
\usepackage{subcaption}
\usepackage{blkarray}

\begin{document}

\title[Impact of Data Distribution on Q-learning with Function Approximation]{The Impact of Data Distribution on Q-learning with Function Approximation}

%%=============================================================%%
%% Prefix	-> \pfx{Dr}
%% GivenName	-> \fnm{Joergen W.}
%% Particle	-> \spfx{van der} -> surname prefix
%% FamilyName	-> \sur{Ploeg}
%% Suffix	-> \sfx{IV}
%% NatureName	-> \tanm{Poet Laureate} -> Title after name
%% Degrees	-> \dgr{MSc, PhD}
%% \author*[1,2]{\pfx{Dr} \fnm{Joergen W.} \spfx{van der} \sur{Ploeg} \sfx{IV} \tanm{Poet Laureate} 
%%                 \dgr{MSc, PhD}}\email{iauthor@gmail.com}
%%=============================================================%%

\author*[1,2]{\fnm{Pedro P.} \sur{Santos}}\email{pedro.pinto.santos@tecnico.ulisboa.pt}

\author[1,2]{\fnm{Diogo S.} \sur{Carvalho}}\email{diogo.s.carvalho@tecnico.ulisboa.pt}

\author[1,2]{\fnm{Alberto} \sur{Sardinha}}\email{jose.alberto.sardinha@tecnico.ulisboa.pt}

\author[1,2]{\fnm{Francisco S.} \sur{Melo}}\email{fmelo@inesc-id.pt}

\affil[1]{\orgname{Instituto Superior Técnico}, \city{Lisbon}, \country{Portugal}}

\affil[2]{\orgname{INESC-ID}, \city{Lisbon}, \country{Portugal}}

%\affil[2]{\orgdiv{Department}, \orgname{Organization}, \orgaddress{\street{Street}, \city{City}, \postcode{10587}, \state{State}, \country{Country}}}

%%==================================%%
%% sample for unstructured abstract %%
%%==================================%%

\abstract{We study the interplay between the data distribution and $Q$-learning-based algorithms with function approximation. We provide a unified theoretical and empirical analysis as to how different properties of the data distribution influence the performance of $Q$-learning-based algorithms. We connect different lines of research, as well as validate and extend previous results. We start by reviewing theoretical bounds on the performance of approximate dynamic programming algorithms. We then introduce a novel four-state MDP specifically tailored to highlight the impact of the data distribution in the performance of $Q$-learning-based algorithms with function approximation, both online and offline. Finally, we experimentally assess the impact of the data distribution properties on the performance of two offline $Q$-learning-based algorithms under different environments. According to our results: (i) high entropy data distributions are well-suited for learning in an offline manner; and (ii) a certain degree of data diversity (data coverage) and data quality (closeness to optimal policy) are jointly desirable for offline learning.}


%%\pacs[JEL Classification]{D8, H51}

%%\pacs[MSC Classification]{35A01, 65L10, 65L12, 65L20, 65L70}

\maketitle

\section{Introduction}
Recent years witnessed significant progress in solving challenging problems across various domains using reinforcement learning (RL) \citep{mnih_2015,silver_2017,lillicrap_2016}. $Q$-learning algorithms with function approximation are among the most used methods \citep{arulkumaran2017brief}. However, the combination of $Q$-learning with function approximation is non-trivial, especially for the case of large capacity approximators such as neural networks. Several works analyze the unstable behavior of such algorithms both experimentally \citep{hasselt_2018,fu_2019} and theoretically \citep{zhang_2021,carvalho2020new}.

The interplay between the data distribution and the outcome of the learning process is one potential source of instability of $Q$-learning-based algorithms \citep{sutton_2018,kumar_2020}. Different lines of research shed some light on how the data distribution impacts algorithmic stability. For example, some works provide examples that induce unstable behavior in off-policy learning \citep{baird_1995,kolter_2011}; some theoretical works derive error bounds on the performance of $Q$-learning-related algorithms \citep{munos_2005,munos_2008,chen_2019}; yet other studies investigate the stability of RL methods with large capacity approximators \citep{fu_2019,kumar_2020}.

We center our study around the following research question: \emph{which data distributions lead to improved algorithmic stability and performance?} In the context of this work, we refer to the data distribution as the distribution used to sample experience or the distribution induced by a dataset of transitions. We investigate how different data distribution properties influence performance in the context of $Q$-learning-based algorithms with function approximation. We add to previous works by providing a systematic and comprehensive study that connects different lines of research, as well as validating and extending previous results, both theoretically and empirically. We primarily focus on offline RL settings with discrete action spaces \citep{levine_2020}, in which an RL agent aims to learn reward-maximizing behavior using previously collected data without additional interaction with the environment. Nevertheless, our conclusions are also relevant in online RL settings, particularly for algorithms that rely on large-scale replay buffers. Our conclusions contribute to a deeper understanding of the influence of the data distribution properties in the performance of approximate dynamic programming (ADP) methods.

We start by presenting some background and the notation used throughout the paper in Sec.~\ref{sec:background}. We connect our work with previous lines of research in Sec.~\ref{sec:related_work}. Then, we investigate how the data distribution impacts the performance of ADP methods. In Sec.~\ref{sec:the_data_distribution_matters:theoretical_bounds}, we review bounds on the performance of ADP methods; we highlight the close relationship between different properties of the data distribution and the tightness of the bounds and motivate high entropy distributions from a game-theoretical point of view. In Sec.~\ref{sec:the_data_distribution_matters:four_state_MDP}, we propose a novel four-state MDP specifically tailored to highlight how the data distribution impacts algorithmic performance, both online and offline. Finally, in Sec.~\ref{sec:experimental_results}, we empirically assess the impact of the data distribution on the performance of offline $Q$-learning-based algorithms with function approximation under different environments, connecting the obtained results with the discussion presented in Sec~\ref{sec:the_data_distribution_matters}. According to our results: (i) high entropy data distributions are well-suited for offline learning; and (ii) a certain degree of data diversity (data coverage) and data quality (closeness to optimal policy) are jointly desirable for offline learning. The results in Sec.~\ref{sec:experimental_results} are one of the main contributions of the paper. Our conclusions appear in Sec.~\ref{sec:conclusion}.

\section{Background}
\label{sec:background}
In RL \citep{sutton_2018}, the agent-environment interaction is modeled as an MDP, formally defined as a tuple $(\mathcal{S}, \mathcal{A}, p, p_0, r, \gamma)$, where $\mathcal{S}$ denotes the state space, $\mathcal{A}$ denotes the action space, $p: \mathcal{S} \times \mathcal{A} \rightarrow \mathcal{P}(\mathcal{S})$ is the state transition probability function with $\mathcal{P}(\mathcal{S})$ being the set of distributions on $\mathcal{S}$, $p_0 \in \mathcal{P}(\mathcal{S})$ is the initial state distribution, $r: \mathcal{S} \times \mathcal{A} \rightarrow \mathbb{R}$ is the reward function, and $\gamma \in (0,1)$ is a discount factor. At each step $t$, the agent observes the state of the environment $s_t \in \mathcal{S}$ and chooses an action $a_t \in \mathcal{A}$. Depending on the chosen action, the environment evolves to state $s_{t+1} \in \mathcal{S}$ with probability $p(s_{t+1} \rvert s_t, a_t)$, and the agent receives a reward $r_t$ with expectation given by $r(s_t,a_t)$. A policy $\pi$ is a mapping  $\pi: \mathcal{S} \rightarrow \mathcal{P}(\mathcal{A})$ encoding the preferences of the agent. We denote by $P^\pi$ the $ \lvert \mathcal{S} \rvert \times \lvert \mathcal{S} \rvert$ matrix with elements $P^\pi(s_t,s_{t+1}) = \mathbb{E}_{a \sim \pi(a \rvert s_t)}\left[ p(s_{t+1} \rvert s_t, a) \right]$. A trajectory, $\tau = (s_0,a_0,...,s_\infty,a_\infty)$, comprises a sequence of states and actions. The probability of a trajectory $\tau$ under a given policy $\pi$ is given by
$$
\varrho_\pi(\tau) = p_0(s_0) \prod_{t=0}^\infty \pi(a_t \rvert s_t) p(s_{t+1} \rvert s_t,a_t).
$$
The discounted reward objective can be written as
$$
J(\pi) = \mathbb{E}_{\tau \sim \varrho_\pi} \left[ \sum_{t=0}^\infty \gamma^t r(s_t,a_t)\right].
$$
The objective of the agent is to find an optimal policy $\pi^*$ that maximizes the objective function above such that $J(\pi^*) \ge J(\pi), \forall \pi$. RL algorithms usually involve the estimation of the optimal action-value function, $Q^*$, satisfying the Bellman optimality equation:
\begin{equation}
\label{eq:bellman_opt_eq}
Q^*(s_t,a_t) = r(s_t,a_t) + \gamma \mathbb{E}_{s_{t+1} \sim p(\cdot \rvert s_t,a_t)}\left[\max_{a' \in \mathcal{A}}Q^*(s_{t+1},a')\right].
\end{equation}
The optimal value function, $V^*$, can be computed from $Q^*$ as $V^*(s_t) = \max_{a_t \in \mathcal{A}} Q^*(s_t,a_t)$. $Q$-learning \citep{watkins_1992} allows an agent to learn directly from raw experience in an online, incremental fashion, by estimating optimal $Q$-values from observed trajectories using the temporal difference (TD) update rule:
$$
Q(s_t,a_t) \leftarrow Q(s_t,a_t) + \alpha \left[ r_t + \gamma \max_{a' \in \mathcal{A}} Q(s_{t+1}, a') - Q(s_t,a_t)\right].
$$
Convergence to the optimal policy is guaranteed if all state-action
pairs are visited infinitely often and the learning rate $\alpha$ is appropriately decayed. Usually, in order to adequately explore the $\mathcal{S}\times \mathcal{A}$ space, an $\epsilon$-greedy policy is used. An $\epsilon$-greedy policy chooses, most of the time, an action that has maximal $Q$-value for the current state, but with probability $\epsilon$ selects a random action instead.

In ADP, $Q$-values are approximated by a differentiable function $Q_\phi$, where $\phi$ denotes the learnable parameters of the model. ADP algorithms, such as the well-known deep $Q$-network (DQN) algorithm \citep{mnih_2015}, usually interleave two phases: (i) a sampling phase, where parameters $\phi$ are kept fixed and a behavior policy (e.g., $\epsilon$-greedy policy) is used to collect transitions $(s_t,a_t,r_t,s_{t+1})$ and store them into a replay buffer, denoted by $\mathcal{B}$; and (ii) an update phase, where transitions are sampled from $\mathcal{B}$ and used to update parameters $\phi$ such that
$$
\mathcal{L}(\phi) = \mathbb{E}_{(s_t,a_t,r_t,s_{t+1}) \sim \mathcal{B}} [ (r_t + \gamma\max_{a' \in \mathcal{A}}Q_{\phi^-}(s_{t+1},a')-Q_\phi(s_t,a_t))^2],
$$
\noindent is minimized, where $\phi^-$ denotes the parameters of the target network, $Q_{\phi^-}$, a periodic copy of the behavior network. The pseudocode of a generic $Q$-learning algorithm with function approximation can be found in Appendix \ref{sec:supplementarysec2}.

Offline RL \citep{levine_2020} aims at finding an optimal policy with respect to $J(\pi)$ using a static dataset of experience. The fundamental problem of offline RL is distributional shift: out-of-distribution samples lead to algorithmic instabilities and performance loss, both at training and deployment time. The conservative $Q$-learning (CQL) \citep{kumar_2020_cql} algorithm is an offline RL algorithm that aims to estimate the optimal $Q$-function using ADP techniques, while mitigating the impact of distributional shift. Precisely, the algorithm avoids the overestimation of out-of-distribution actions by considering an additional conservative penalty term of the type $\mathcal{L}_c=\mathbb{E}_{s\sim\mathcal{B},a\sim\nu}[Q_\phi(s,a)]$, yielding the global objective $\mathcal{L}_{\text{CQL}} = \mathcal{L}_\phi + k \mathcal{L}_c, k \in \mathbb{R}_0^+$, which the algorithm aims to minimize. Distribution $\nu(a \rvert s)$ is chosen adversarially such that it selects overestimated $Q$-values with high probability, e.g., by maximizing $\mathcal{L}_c$.

\section{Related Work}
\label{sec:related_work}
We now review and discuss several lines of research that, in one way or another, are related to the problem of studying the impact of data distribution on $Q$-learning-related algorithms. We organize our discussion around three main topics: (i) studies that derive error bounds on the performance of $Q$-learning-related algorithms (Sec.~\ref{sec:related_work:error_propagation}); works that analyze the unstable behavior in off-policy learning (Sec.~\ref{sec:related_work:unstable_behavior}); and (iii) studies that investigate the stability of deep RL methods and propose algorithms for offline RL (Sec.~\ref{sec:related_work:stability_deep_rl}).

\subsection{Error propagation in ADP}
\label{sec:related_work:error_propagation}
On a theoretical side, there is a number of different works that analyze error propagation in ADP methods, deriving error bounds on the performance of approximate policy iteration \cite{kakade_2002,munos_2003} and approximate value iteration \citep{munos_2005,munos_2008} algorithms. \citet{munos_2003} provides error bounds for approximate policy iteration using quadratic norms, as well as bounds on the error between the performance of the policies induced by the value iteration algorithm and the optimal policy as a function of weighted $L_p$-norms of the approximation errors \citep{munos_2005}. \citet{munos_2008} develop a theoretical analysis of the performance of sampling-based fitted value iteration, providing finite-time bounds on the performance of the algorithm. %\citet{farahmand_2010} study how the approximation error/Bellman residual at each iteration of the approximate policy/value iteration algorithms influences the quality of the resulting policies.
\citet{yang_2019} establish algorithmic and statistical rates of convergence for the iterative sequence of $Q$-functions obtained by the DQN algorithm. \citet{chen_2019} further improve the bounds of the previous studies while considering an offline RL setting. Common to all these works is the dependence of the derived bound on concentrability coefficients that heavily depend on the data distribution. In this work, we review concentrability coefficients in Sec. \ref{sec:the_data_distribution_matters:theoretical_bounds}, providing a motivation for the use of high entropy data distributions through the lens of robust optimization. We also analyze our experimental results, presented in Sec. \ref{sec:experimental_results}, in light of the theoretical results from these previous articles.

\subsection{Unstable behavior in off-policy learning}
\label{sec:related_work:unstable_behavior}
Several early studies analyze the unstable behavior of off-policy learning algorithms and the harmful learning dynamics that can lead to the divergence of the function parameters \citep{kolter_2011,tsitsiklis_1996,tsitsiklis_1997,baird_1995}. For instance, \citet{baird_1995,tsitsiklis_1996,kolter_2011} provide examples that highlight the unstable behavior of ADP methods. \citet{kolter_2011} provides an example that highlights the dependence of the off-policy distribution on the approximation error of the algorithm. In Sec.~\ref{sec:the_data_distribution_matters:four_state_MDP}, we propose a novel four-state MDP that highlights the impact of the data distribution in the performance of ADP methods. We further explore how off-policy algorithms are affected by data distribution changes, under diverse settings. We add to previous works by considering both offline settings comprising static data distributions, and online settings in which data distributions are induced by a replay buffer.

\subsection{The stability of deep and offline RL algorithms}
\label{sec:related_work:stability_deep_rl}
Several works investigate the stability of deep RL methods \citep{hasselt_2018,fu_2019,kumar_2019,kumar_2020,liu_2018,zhang_2021,wang_2021}, as well as the development of RL methods specifically suited for offline settings \citep{agarwal_2019,mandlekar_2021,levine_2020}. For example, \citet{kumar_2020} observe that $Q$-learning-related methods can exhibit pathological interactions between the data distribution and the policy being learned, leading to potential instability. \citet{fu_2019} investigate how different components of DQN play a role in the emergence of the deadly triad. In particular, the authors assess the performance of DQN with different sampling distributions, finding that higher entropy distributions tend to perform better. \citet{agarwal_2019} provide a set of ablation studies that highlight the impact of the dataset size and diversity in offline learning settings. \citet{wang_2021} study the stability of offline policy evaluation, showing that even under relatively mild distribution shift, substantial error amplification can occur. In Sec.~\ref{sec:experimental_results}, we provide a systematic study on how different properties of the data distribution impact the performance of deep offline RL algorithms by directly controlling the dataset generation process, allowing us to rigorously control different datasets' metrics and systematically compare our experimental results. We validate and extend previous results, as well as discuss our experimental findings in light of existing theoretical results.

\bigbreak

Finally, \citet{schweighofer_2021} study the impact of dataset characteristics on offline RL. More exactly, the authors study the influence of the average dataset return and state-action coverage on the performance of different RL algorithms while controlling the dataset generation procedure. Despite some similarities with the experiments in Sec. \ref{sec:experimental_results}, we present a much broader picture regarding the impact of the data distribution on the stability of general off-policy RL algorithms, from both theoretical and experimental point-of-views. Additionally, the experimental methodology carried out by both works differs in several aspects, such as the calculation of the dataset metrics used for the presentation of the experimental results or the types of environments used.

\section{Data distribution matters}
\label{sec:the_data_distribution_matters}
In this section, we show that the data distribution plays an important role in regulating the performance of $Q$-learning-based algorithms, offering theoretical and empirical evidence to support this claim. In Sec.~\ref{sec:the_data_distribution_matters:theoretical_bounds}, we review and analyze the role played by concentrability coefficients in upper error bounds of ADP methods. Then, in Sec.~\ref{sec:the_data_distribution_matters:four_state_MDP}, we propose a four-state MDP designed to highlight the impact of the data distribution on the performance of ADP methods.

\subsection{Concentrability coefficients}
\label{sec:the_data_distribution_matters:theoretical_bounds}
As discussed in Sec.~\ref{sec:related_work:error_propagation}, several works analyze error propagation in ADP \citep{munos_2003,munos_2005,munos_2008,yang_2019,chen_2019}. Specifically, the aforementioned works provide upper bounds of the type $\| V^* - V^{\pi_k}\|_{p,\rho} \le C \cdot \mathcal{F} + \mathcal{E}$ or $\| Q^* - Q^{\pi_k} \|_{p,\rho} \le C \cdot \mathcal{F} + \mathcal{E}$. Intuitively, the bounds correspond to $\rho$-weighted $L_p$-norms $(p \ge 1)$ between $V^*$/$Q^*$, and the value/action-value function induced by the greedy policy $\pi_k$ with respect to the estimated value/action-value function at the $k$-th timestep\footnote{We use the bounds of \citet{munos_2008} as reference.}. Such bounds comprise, in general, three key components:
\begin{enumerate}
    \item A concentrability coefficient, $C$, that quantifies the suitability of the sampling distribution $\mu \in \mathcal{P}(\mathcal{S})$ or $\mu \in \mathcal{P}(\mathcal{S}, \mathcal{A})$.
    \item A measure of the approximation power of the function space, $\mathcal{F}$, which reflects how well the function space is aligned with the dynamics and reward of the MDP.
    \item A coefficient $\mathcal{E}$ that captures the sampling error of the algorithm, i.e., the error that accumulates due to limited sampling and iterations.  
\end{enumerate}
From the three components above, we focus our attention on the study of the concentrability coefficient as it captures the impact of the data distribution in the tightness of the upper bound.

\citet{munos_2003} introduces the first version of this data-dependent concentrability coefficient, which is related to the density of the transition probability function. Specifically, the author defines the coefficient $C_1 \in \mathbb{R}^+ \cup \{+\infty\}$ as
\begin{equation}
C_1 = \sup_{s, s' \in \mathcal{S}, a \in \mathcal{A}} \frac{p(s' \rvert s,a)}{\mu(s')},
\label{eq:C1_bound}
\end{equation}
\noindent with $\mu \in \mathcal{P}(\mathcal{S})$ and the convention that $0/0=0$, and $C_1=\infty$ if $\mu(s') = 0$ and $p(s' \rvert s,a) > 0$. We use this convention for all upcoming coefficients. Intuitively, the noisier the dynamics of the MDP, the smaller the coefficient $C_1$ and the tighter the bound.
\citet{munos_2005} introduces a different concentrability coefficient related to the discounted average concentrability of future-states on the MDP. Specifically, coefficient $C_2 \in \mathbb{R}^+ \cup \{+\infty\}$ is defined as
\begin{equation}
\label{eq:C2_bound}
C_2 = (1-\gamma)^2\sum_{m = 1}^{\infty} m \gamma^{m-1} c(m),
\end{equation}
\begin{equation}
\label{eq:supremum_coef}
c(m) = \sup_{\pi_1, ..., \pi_m \in \Pi, s \in \mathcal{S}}\frac{(\rho P^{\pi_1} P^{\pi_2} ... P^{\pi_m})(s)}{\mu(s)},
\end{equation}
with $\mu, \rho \in \mathcal{P}(\mathcal{S})$, $\Pi$ denotes the space of all possible policies, and $\rho$ reflects the importance of various regions of the state space and is selected by the practitioner. Intuitively, coefficient $C_2$ expresses some smoothness property of the future state distribution with respect to $\mu$ for an initial distribution $\rho$. \citet{munos_2005} and \citet{munos_2008} note that the assumption that $C_1 < \infty$ is stronger than the assumption that $C_2 < \infty$.

\citet{farahmand_2010} and \citet{yang_2019} replace the supremum norm of \eqref{eq:supremum_coef} with a weighted norm of the type
\begin{equation}
\label{eq:expectation_coef}
c(m) = \sup_{\pi_1, ..., \pi_m \in \Pi} 
 \left( \mathbb{E}_{(s,a) \sim \mu} \left[ \left\lvert\frac{(\rho P^{\pi_1} P^{\pi_2} ... P^{\pi_m})(s,a)}{\mu(s,a)} \right\rvert^2 \right] \right)^{1/2},
\end{equation}

\noindent with $\mu, \rho \in \mathcal{P}(\mathcal{S} \times \mathcal{A})$. We let $C_3 \in \mathbb{R}^+ \cup \{+\infty\}$ denote the coefficient defined by \eqref{eq:C2_bound} and \eqref{eq:expectation_coef}.

Other works \citep{munos_2007,chen_2019,antos_2008,lazaric_2012,lazaric_2016,tosatto_2017,xie_2020} use concentrability coefficients similar to those defined above to derive upper bounds on the performance of various algorithms.

Unfortunately, although the concentrability coefficients above attempt to quantify distributional shift, they have limited interpretability. Specifically, it is hard to infer from the coefficients above which exact sampling distributions should be used. For example, if we consider coefficients $C_2$ and $C_3$, even if we know which parts of the state space are relevant according to the distribution $\rho$, the computation of the coefficient in \eqref{eq:C2_bound} still depends on the complex interactions between $\rho$ and the dynamics of the MDP under any possible policy. What can be concluded is that the concentrability coefficient will depend on all states that can be reached by any policy when the starting state distribution is given by $\rho$. However, it is not obvious which exact target distribution we should aim at when selecting the sampling distribution $\mu$, especially when we do not have access to the full specification of the underlying MDP or which regions of the state space are of interest. In the face of such uncertainty, previous works assume sufficient coverage of the state (and action) space, thus using upper bounded concentrability coefficients to analyze the performance of the algorithms \citep{munos_2008,farahmand_2010,chen_2019,yang_2019}.

More recently, \citet{chen_2019} revisit the assumption of a bounded concentrability coefficient and formally justify the necessity of mild distribution shift via an information-theoretic lower bound. Precisely, the authors show that, under a bounded concentrability coefficient defined using \eqref{eq:supremum_coef}, polynomial sample complexity is precluded if the MDP dynamics are not restricted. In subsequent work, \citet{xie_2020} break the hardness conjecture introduced by \citet{chen_2019} albeit under a more restrictive concentrability coefficient similar to that of \eqref{eq:C1_bound}. Both works highlight the dependence of the concentrability coefficient on the properties of the MDP.

Other works \citep{wang_2020,amortila_2020,zanette_2020} have also proved hardness results for offline RL, however, under an even weaker form of concentrability than that induced by Eqs.~\eqref{eq:C1_bound} and \eqref{eq:supremum_coef}. For example, \citet{wang_2020} show that good coverage over the feature space is not sufficient to sample-efficiently perform offline policy evaluation with linear function approximation and that significantly stronger assumptions on distributional shift may be needed. We refer to \citet{xie_2020} for a detailed discussion on the relation between the different proposed concentrability coefficients and hardness results for offline RL.

We conclude by noting that offline policy evaluation allows for more interpretable and amenable quantification of distributional shift; under such setting, distributional shift can be more accurately quantified in terms of a statistical distance between sampling distribution $\mu$ and the stationary distribution of the target policy, as recently proposed by \citet{duan_2020}.

In the next section, we provide an interpretation of $C_3$ as an $f$-divergence and give a new motivation for the use of maximum entropy sampling distributions from a game-theoretical point of view.

\subsubsection{Motivating Maximum Entropy Distributions}
\label{sec:the_data_distribution_matters:theoretical_bounds:max_ent_motivation}
Letting $\beta=\rho P^{\pi_1} P^{\pi_2} ... P^{\pi_m}$, we can rewrite \eqref{eq:expectation_coef} as
\begin{align*}
 \left\| \frac{\beta}{\mu}\right\|_{2,\mu} &= \left(\mathbb{E}_{(s,a)\sim \mu}\left[\left(\frac{\beta(s,a)}{\mu(s,a)}\right)^2\right]\right)^{1/2} =\sqrt{\mathcal{D}_{f}(\beta \rvert\rvert \mu) + 1},
\end{align*}
\noindent for $f(x)=x^2 - 1$, where $\mathcal{D}_f$ denotes the $f$-divergence. Optimizing \eqref{eq:expectation_coef} over the distribution $\mu$ is hard due to the fact that we actually want to minimize $\mathcal{D}_{f}(\beta \rvert\rvert \mu)$ with respect to a large set of different $\beta$ distributions, due to the supremum in \eqref{eq:expectation_coef}, as well as the summation in \eqref{eq:C2_bound}. Furthermore, we usually do not know the transition probability function. Therefore, we analyze the problem of picking an optimal $\mu$ distribution as a robust optimization problem. Specifically, we formulate a minimax objective where the minimizing player aims at choosing $\mu$ to minimize $\mathcal{D}_{f}(\beta \rvert\rvert \mu)$ and the maximizing player chooses $\beta$ to maximize $\mathcal{D}_{f}(\beta \rvert\rvert \mu)$.
\begin{proposition}
\label{theo:minimax}
Let $\mathcal{P}(\mathcal{S} \times \mathcal{A})$ represent the set of probability distributions over $\mathcal{S} \times \mathcal{A}$. Let also $L_\mu: \mathcal{P}(\mathcal{S} \times \mathcal{A}) \to \mathbb{R}^+_0$ such that $L_\mu(\beta) = \left\| \frac{\beta}{\mu} \right\|_{2, \mu}^2$.
The solution $\mu$ to %the minimax objective
\begin{equation}\label{game:minimax}
    \argmin_{\mu \in \mathcal{P}(\mathcal{S} \times \mathcal{A})} \max_{\beta \in \mathcal{P}(\mathcal{S} \times \mathcal{A})} L_\mu(\beta)
\end{equation}
is the maximum entropy distribution over $\mathcal{S} \times \mathcal{A}$. Proof in Appendix \ref{sec:supplementarysec4}.
\end{proposition}
As stated in Proposition \ref{theo:minimax}, the maximum entropy distribution is the solution to the robust optimization problem. This result provides a theoretical justification for the benefits of using high entropy sampling distributions, as suggested by previous works \citep{kakade_2002,munos_2003}: in the face of uncertainty regarding the underlying MDP, high entropy distributions ensure coverage over the state-action space, thus contributing to keep concentrability coefficients bounded.

\bigbreak 

The bounds surveyed in this section suggest that high coverage over the state-action space and high entropy distributions are beneficial. However, it is important to note that the significance of the previous results is highly dependent on the actual tightness of the bound \citep{munos_2005}; rather loose bounds can trivially upper bound the error but be of little help to understanding algorithmic behavior. Thus, it is important to understand, from a practical point-of-view, if the properties suggested by the surveyed bounds contribute to improved algorithmic performance. Therefore, we investigate, from an experimental point-of-view, how the data distribution impacts performance: (i) in the next section (Sec.~\ref{sec:the_data_distribution_matters:four_state_MDP}) under our four-state MDP; and (ii) in Sec.~\ref{sec:experimental_results} under high dimensional environments and two $Q$-learning-based algorithms.

\subsection{Four-state MDP}
\label{sec:the_data_distribution_matters:four_state_MDP}
\begin{figure}[t]
\centering
\includegraphics[width=0.5\textwidth]{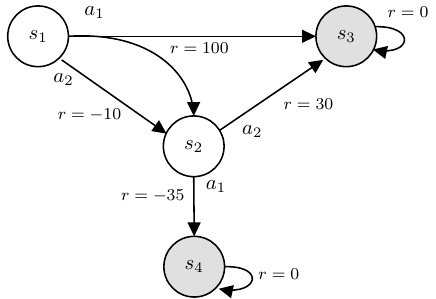}
\caption{Four-state MDP, with states $\{s_1,s_2,s_3,s_4\}$ and actions $\{a_1,a_2\}$. State $s_1$ is the initial state and states $s_3$ and $s_4$ are terminal (absorbing) states. All actions are deterministic except for the state-action pair $(s_1,a_1)$, where $p(s_3 \rvert s_1,a_1) = 0.99$ and $p(s_2 \rvert s_1,a_1) = 0.01$. The reward function is $r(s_1,a_1) = 100$, $r(s_1,a_2) = -10$, $r(s_2,a_1) = -35$ and $r(s_2,a_2) = 30$.}
\label{fig:four_states_mdp}
\end{figure}

We now study how the data distribution influences the performance of a $Q$-learning algorithm with function approximation under the four-state MDP (Fig. \ref{fig:four_states_mdp}). We show that the data distribution can significantly influence the quality of the resulting policies and affect the stability of the learning algorithm. Due to space constraints, we focus our discussion on the main conclusions and refer to Appendix~\ref{sec:supplementarysec4} for an in-depth discussion.

We focus our attention to non-terminal states $s_1$ and $s_2$ and set $\gamma = 1$. In state $s_1$ the correct action is $a_1$, whereas in state $s_2$ the correct action is $a_2$. We consider a linear function approximator $Q_w(s_t,a_t) = w^\mathrm{T} \phi(s_t,a_t)$, where $\phi$ is a feature mapping, defined as $\phi(s_1,a_1) = [1,0,0]^\mathrm{T}$, $\phi(s_1,a_2) = [0,1,0]^\mathrm{T}$, $\phi(s_2,a_1) = [\alpha,0,0]^\mathrm{T}$, and $\phi(s_2,a_2) = [0,0,1]^\mathrm{T}$, with $\alpha \in [1, 3/2)$. As can be seen, %due to the choice of feature mapping,  
the capacity of the function approximator is limited and there exists a correlation between $Q_w(s_1,a_1)$ and $Q_w(s_2,a_1)$. This will be key to the results that follow.

\subsubsection{Offline Learning}
\begin{figure}[t]
\begin{center}
    \includegraphics[width=0.5\textwidth]{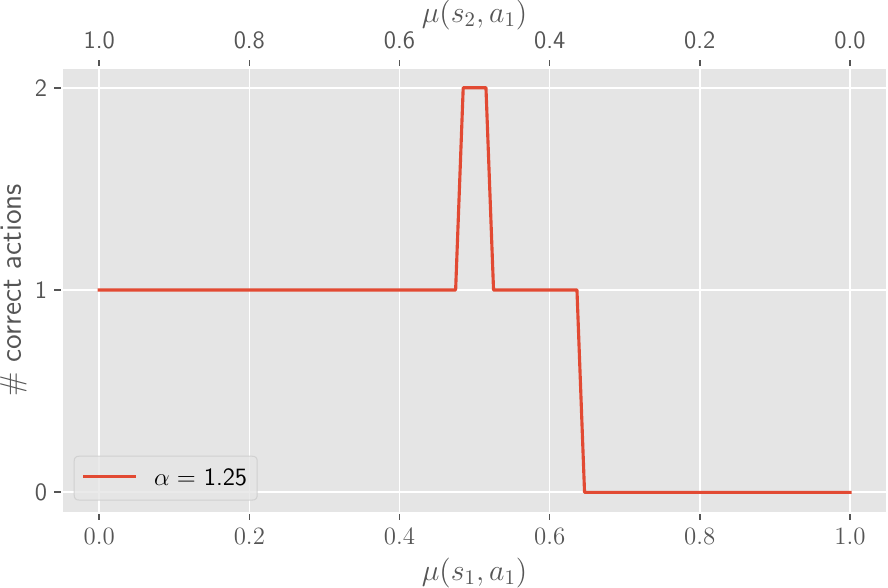}
\caption{The number of correct actions at states $s_1$ and $s_2$ for different data distributions ($\alpha = 1.25$).}
\label{fig:main_correct_actions_TD_target}
\end{center}
\end{figure}

We consider an offline RL setting and denote by $\mu$ the distribution over $\mathcal{S}\times \mathcal{A}$ induced by a static dataset of transitions. We focus our attention on probabilities $\mu(s_1,a_1)$ and $\mu(s_2,a_1)$, since these are the probabilities associated with the two partially correlated state-action pairs.
Fig. \ref{fig:main_correct_actions_TD_target} displays the influence of the data distribution, namely the proportion between $\mu(s_1,a_1)$ and $\mu(s_2,a_1)$, on the number of correct actions yielded by the learned policy. We identify three regimes: (i) when $\mu(s_1,a_1) \approx 0.5$, we learn the optimal policy; (ii) if $\mu(s_1,a_1) < (\approx0.48)$ or $ (\approx 0.52) < \mu(s_1,a_1) < (\approx 0.65)$, the policy is only correct at one of the states; (iii) if $\mu(s_1,a_1) > (\approx 0.65)$, the policy is wrong at both states. The results above show that, due to the limited power and correlation between features, the data distribution impacts performance as the number of correct actions is directly dependent on the properties of the data distribution. As our results show, due to bootstrapping, it is possible that under certain data distributions neither action is correct.

\subsubsection{Online Learning with Unlimited Replay}

\begin{figure}[t]
\begin{center}
\begin{tabular}{cc}
  \includegraphics[width=0.42\textwidth]{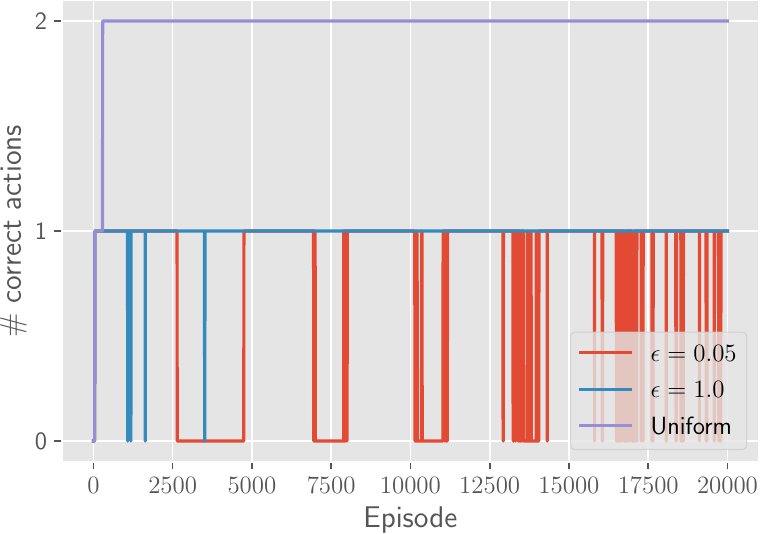} & \includegraphics[width=0.44\textwidth]{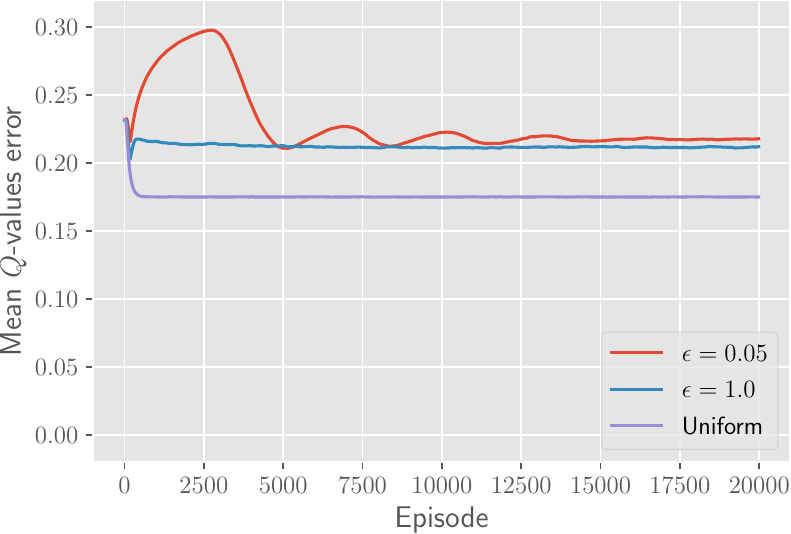} \\
  (a) Number of correct actions. & (b) $Q$-values mean error. \\
\end{tabular}
\caption{Experiments for different exploratory policies with an infinitely-sized replay buffer.}
\label{fig:main_four_states_mdp_exps}
\end{center}
\end{figure}

Instead of considering a fixed $\mu$ distribution, we now consider a setting where $\mu$ is dynamically induced by a replay buffer obtained using $\epsilon$-greedy exploration. Figure~\ref{fig:main_four_states_mdp_exps} shows the results when $\alpha = 1.2$, under: (i) an $\epsilon$-greedy policy with $\epsilon=1.0$; and (ii) an $\epsilon$-greedy policy with $\epsilon=0.05$. We consider a replay buffer with unlimited capacity. We use a uniform data distribution as baseline\footnote{We note that the uniform distribution over the state-action space may be outside the space of possible distributions that can be generated by running policies on the MDP.}. As seen in Fig. \ref{fig:main_four_states_mdp_exps}, the baseline outperforms all other data distributions, as expected given our discussion in the previous section. Regarding the $\epsilon$-greedy policy with $\epsilon=1.0$, the agent is only able to pick the correct action at state $s_1$, featuring a higher average $Q$-value error in comparison to the baseline. This is due to the fact that the stationary distribution of the MDP under the fully exploratory policy is too far from the uniform distribution to retrieve the optimal policy. Finally, for the $\epsilon$-greedy policy with $\epsilon=0.05$, the performance of the agent further deteriorates. Such exploratory policy induces oscillations in the $Q$-values (Fig. \ref{fig:main_four_states_mdp_exps} (b)), which eventually damp out as learning progresses. The oscillations are due to an undesirable interplay between the features %of the function approximator 
and the data distribution: exploitation may cause abrupt changes in the data distribution and hinder learning.

\subsubsection{Online Learning with Limited Replay Capacity}
\begin{figure}[t]
\begin{center}
    \includegraphics[width=0.5\textwidth]{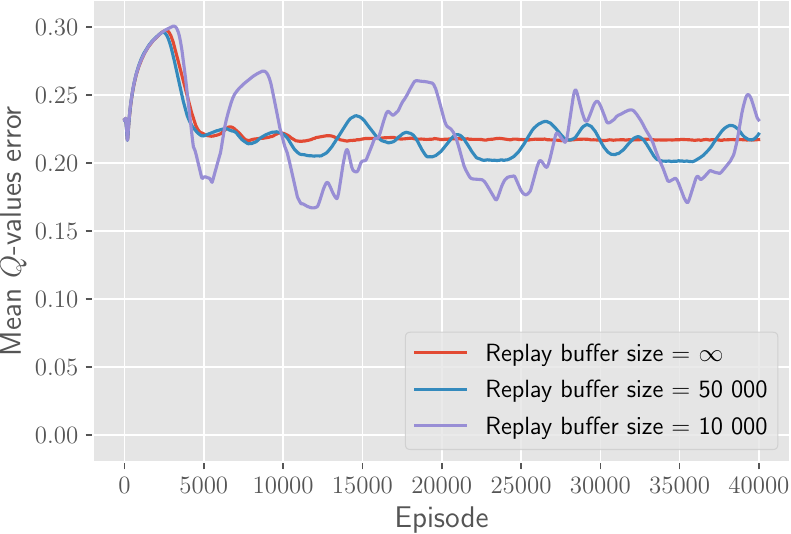}
\caption{$Q$-values error under the $(\epsilon=0.05)$-greedy policy for different replay capacities.}
\label{fig:four_states_mdp_replay_buffer_sizes}
\end{center}
\end{figure}

Finally, we consider an experimental setting where the replay buffer has limited capacity and study the impact of its size in the stability of the algorithm. Figure \ref{fig:four_states_mdp_replay_buffer_sizes} displays the results obtained with the $\epsilon$-greedy policy with $\epsilon=0.05$, while varying the capacity of the replay buffer. As can be seen, as the replay buffer size increases the oscillations in the $Q$-values errors are smaller. The undesirable interplay previously observed under the infinitely-sized replay buffer repeats. However, the smaller the replay buffer capacity, the more the data distribution induced by the contents of the replay buffer is affected by changes to the current exploratory policy, i.e., exploitation leads to more abrupt changes in the data distribution, which, in turn, drive abrupt changes to the $Q$-values. For the infinitely-sized replay buffer the amplitude of the oscillations is dampened because previously stored experience contributes to make the data distribution more stationary, not as easily achieved by smaller buffers.

\subsubsection{Discussion}
We presented a set of experiments using a four-state MDP that shows how the data distribution can influence the performance of the resulting policies and the stability of the learning algorithm. First, we showed that, under an offline RL setting, the number of optimal actions identified is directly dependent on the properties of the data distribution due to an undesirable correlation between features. Second, not only the quality of the computed policies depends on the data collection mechanism, but also an undesirable interplay between the data distribution and the function approximator can arise: exploitation can lead to abrupt changes in the data distribution and hinder learning. Finally, we showed that the replay buffer size can also affect the learning dynamics. Despite the fact that we study a four-state MDP, we argue that the example here presented is still relevant under more realistic settings; we further elaborate on this point in Appendix~\ref{sec:supplementarysec4}.

\section{Assessing the Impact of Data Distribution in Offline RL}
\label{sec:experimental_results}
In this section, we experimentally assess the impact of different data distribution properties on the performance of offline DQN \citep{mnih_2015} and CQL \citep{kumar_2020_cql}. We evaluate the performance of the algorithms under six different environments: the \textit{grid 1} and \textit{grid 2} environments consist of standard tabular environments with highly uncorrelated state features, the \textit{multi-path} environment is a hard exploration environment, and the \textit{pendulum}, \textit{mountaincar} and \textit{cartpole} environments are benchmarking environments featuring a continuous state-space domain. All reported values are calculated by aggregating the results of different training runs. The description of the experimental environments and the experimental methodology, as well as the complete results, can be found in Appendix \ref{sec:supplementarysec5}. The developed software can be found at \url{https://github.com/PPSantos/rl-data-distribution-public}. We also provide an interactive dashboard with all our experimental results at \url{https://rldatadistribution.pythonanywhere.com/}.

In this section, we denote by $\mu$ the data distribution over state-action pairs induced by a static dataset of transitions. We consider two types of offline datasets: (i) \textit{$\epsilon$-greedy} datasets, generated by running an $\epsilon$-optimal policy on the MDP, i.e., a policy that is $\epsilon$-greedy with respect to the optimal $Q$-values, with $\epsilon \in [0,1]$; and (ii) \textit{Boltzmann$(T)$} datasets, generated by running a Boltzmann policy with respect to the optimal $Q$-values with temperature coefficient $T \in [-10,10]$. Additionally, we artificially enforce that some of the generated datasets have full coverage over the $\mathcal{S} \times \mathcal{A}$ space. We do this by running an additional procedure that ensures that each state-action pair appears at least once in the dataset. We chose not to use publicly available datasets for offline RL \citep{fu_2020,gulcehre_2020,qin_2021} in order to have complete control over the dataset generation procedure, which allows us to rigorously control different datasets' metrics and systematically compare our experimental results. Nevertheless, our results are representative of a diverse set of discrete action-space control tasks.

Two aspects are worth highlighting. First, in all environments, the sampling error is low due to the highly deterministic nature of the underlying MDPs. Thus, a single next-state sample is sufficient to correctly evaluate the Bellman optimality operator (Eq.~\eqref{eq:bellman_opt_eq}). Second, the function approximator has enough capacity to correctly represent the optimal $Q$-function, a property known as \emph{realizability} \citep{chen_2019}.

%We break down our analysis in different sections, each covering a different perspective on the impact of the data distribution in the performance of the offline RL algorithms, and terminate with a more high-level and unified discussion.

\subsection{High entropy is beneficial}
\label{sec:experimental_results:entropy}
\begin{figure}[t]
\begin{center}
    \includegraphics[width=0.6\textwidth]{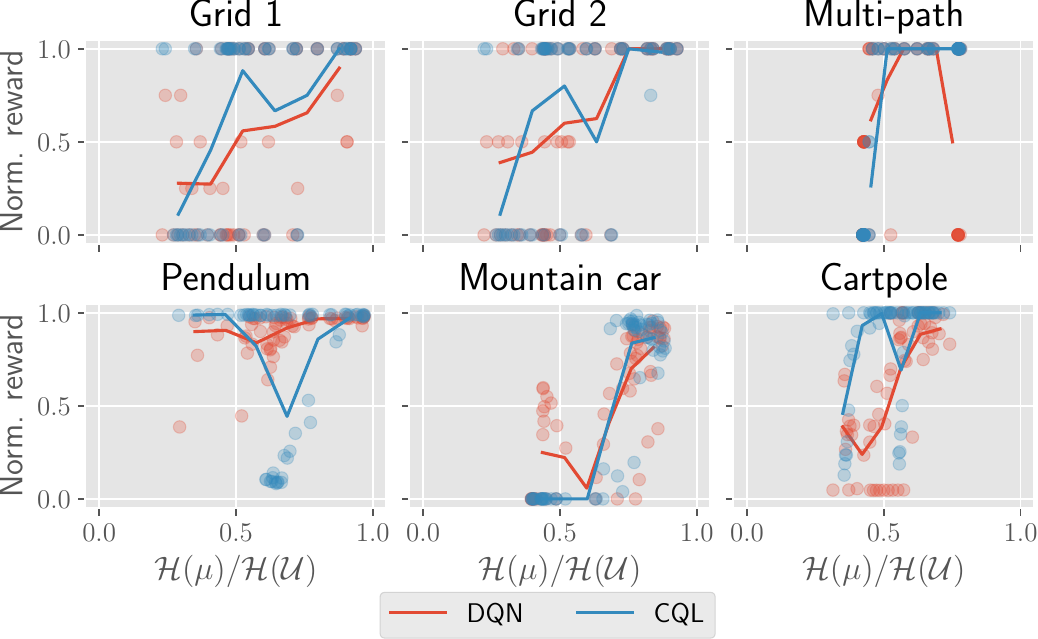}
\caption{Average normalized rollouts reward for datasets with different normalized entropies.}
\label{fig:exp_results:entropy_plot}
\end{center}
\end{figure}

We start our analysis by studying the impact of the dataset distribution entropy, $\mathcal{H}(\mu)$, on the performance of the offline RL algorithms. Figure \ref{fig:exp_results:entropy_plot} displays the average normalized rollouts reward for datasets with different normalized entropies. As can be seen, under all environments and for both offline RL algorithms, high entropy distributions tend to achieve increased rewards. In other words, distributions with a large entropy appear to be well-suited to be used in offline learning settings. Such observation is inline with the discussion in Sec. \ref{sec:the_data_distribution_matters:theoretical_bounds} and works such as \citep{kakade_2002,munos_2003}: high entropy distributions contribute to increased coverage, keeping concentrability coefficients bounded and, thus, mitigating algorithmic instabilities.

Importantly, we do not claim that high entropy distributions are the only distributions suitable to be used. As can be seen in Fig. \ref{fig:exp_results:entropy_plot}, certain lower-entropy distributions also perform well. In the next sections, we investigate which other properties of the distribution are of benefit to offline RL.

\subsection{Dataset coverage matters}
\label{sec:experimental_results:coverage}
We now study the impact of dataset coverage, i.e., the diversity of the transitions in the dataset, in the performance of the offline agents. In order to keep the discussion concise, in this section we focus our attention on \textit{$\epsilon$-greedy} datasets, and refer to Appendix \ref{sec:supplementarysec5} for the complete results.

\begin{figure*}[t]
    \centering
    \begin{subfigure}[b]{0.49\textwidth}
        \centering
        \includegraphics[width=0.99\textwidth]{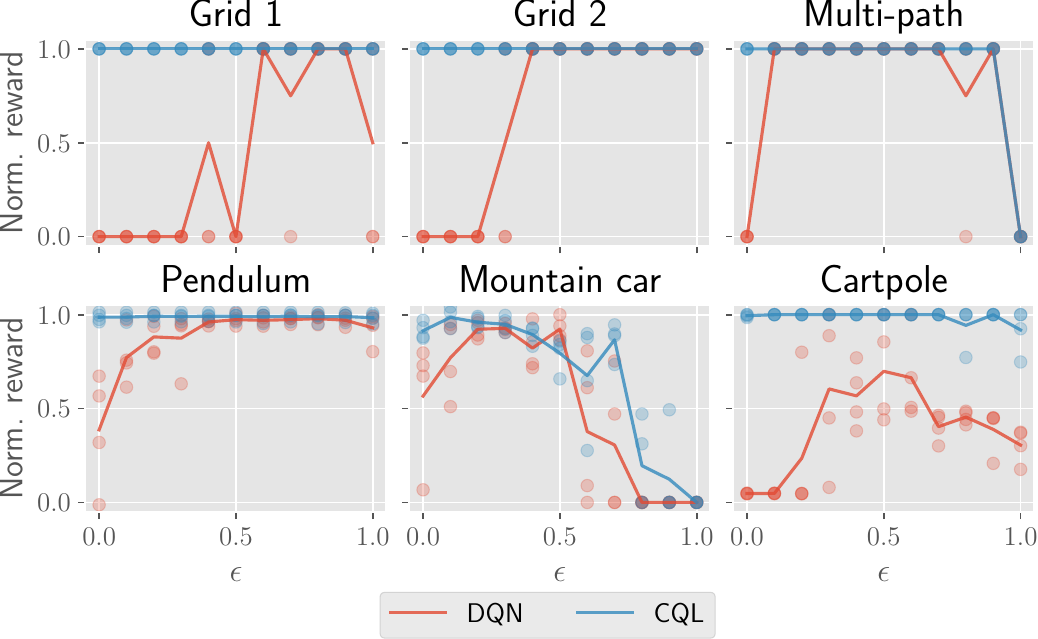}
        \caption{Dataset coverage not enforced.}
    \end{subfigure}
    \hfill
    \begin{subfigure}[b]{0.49\textwidth}
        \centering
        \includegraphics[width=0.99\textwidth]{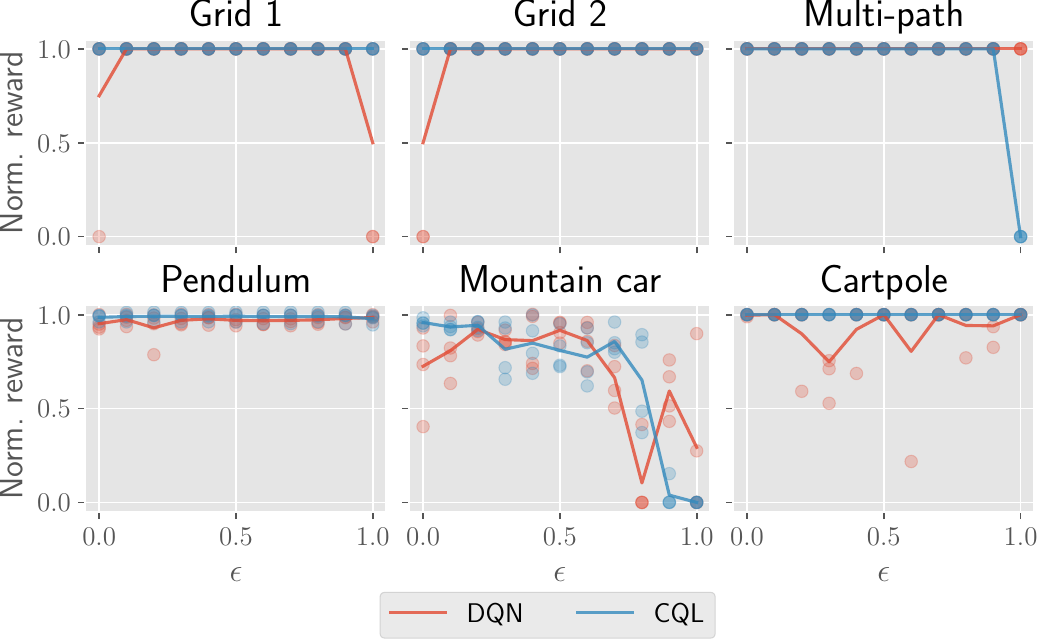}
        \caption{Dataset coverage enforced.}
    \end{subfigure}
    \caption{Average normalized rollouts reward under $\epsilon$-greedy datasets for different $\epsilon$ values.}
    \label{fig:exp_results:coverage}
\end{figure*}

We start by focusing our attention on the offline DQN algorithm. Figure \ref{fig:exp_results:coverage} (a) displays the average normalized rollouts reward under $\epsilon$-greedy datasets with dataset coverage not enforced. As can be seen, DQN struggles to achieve optimal rewards for low values of $\epsilon$, i.e., even though the algorithm is provided with optimal or near-optimal trajectories, it is unable to steadily learn under such setting. However, as $\epsilon$ increases, the performance of the algorithm increases, eventually decaying again for high $\epsilon$ values. Such results suggest that a certain degree of data coverage is required by DQN to robustly learn in an offline manner, despite being provided with high quality data (rich in rewards). On the other hand, the decay in performance for highly exploratory policies under some environments can be explained by the fact that such policies induce trajectories that are poor in reward (this is further explored in the next section). Figure \ref{fig:exp_results:coverage} (b) displays the obtained experimental results under the exact same datasets, except that we enforce coverage over $\mathcal{S} \times \mathcal{A}$. We note a substantial improvement in the performance of DQN across all environments, supporting our hypothesis that data coverage plays an important role regulating the stability of offline RL algorithms.

The CQL algorithm appears to perform more robustly than DQN. Particularly, as seen in Fig. \ref{fig:exp_results:coverage} (a), CQL is able to robustly learn with low $\epsilon$ values, i.e., using optimal or near optimal trajectories that feature low coverage. Additionally, no substantial performance gain is observed under the offline CQL agent by enforcing dataset coverage (Fig. \ref{fig:exp_results:coverage} (b)).

The finding that data coverage appears to play an important role regulating the performance of DQN, even when considering high quality, near optimal trajectories, is inline with the discussion presented in Sec. \ref{sec:the_data_distribution_matters:theoretical_bounds}. One could argue that we are only interested in correctly estimating the $Q$-values along an optimal trajectory, however, due to the bootstrapped nature of the updates, error in the estimation of the $Q$-values for adjacent states can erroneously affect the estimation of the $Q$-values along the optimal trajectory. The argument above is suggested by concentrability coefficients. If we consider distribution $\rho$ from \eqref{eq:supremum_coef} or \eqref{eq:expectation_coef} to be the uniform distribution over the states of the optimal trajectory and zero otherwise, we can see that the concentrability coefficient given by \eqref{eq:C2_bound} still depends on other states than those of the optimal trajectory. Precisely, the coefficient depends on all the states that can be reached by any policy when the starting state is sampled according to $\rho$ (the importance of each state geometrically decays depending on their distance to the optimal trajectory). Therefore, in order to keep the concentrability coefficient low, it is important that such states are present in the dataset. On the other hand, CQL is still able to robustly learn using high quality trajectories independently of the data coverage because of its pessimistic nature. Since the algorithm penalizes the $Q$-values for actions that are underrepresented in the dataset, the error for adjacent states is not propagated in the execution of the algorithm.

\bigbreak

In this section, we considered datasets that are, in general, aligned and close to that induced by optimal policies. What happens if our data is collected by arbitrary policies? We investigate the impact of the trajectory quality in the next section.

\subsection{Closeness to optimal policy matters}
\label{sec:experimental_results:closeness_to_opt_policy}
We now investigate how offline agents are affected by the quality of the trajectories contained in the dataset. More precisely, we study how the statistical distance between distribution $\mu$ and the distribution induced by one of the optimal policies of the MDP, $d_{\pi^*}$, affects offline learning.

\begin{figure*}[t]
    \centering
    \begin{subfigure}[b]{0.49\textwidth}
        \centering
        \includegraphics[width=0.99\textwidth]{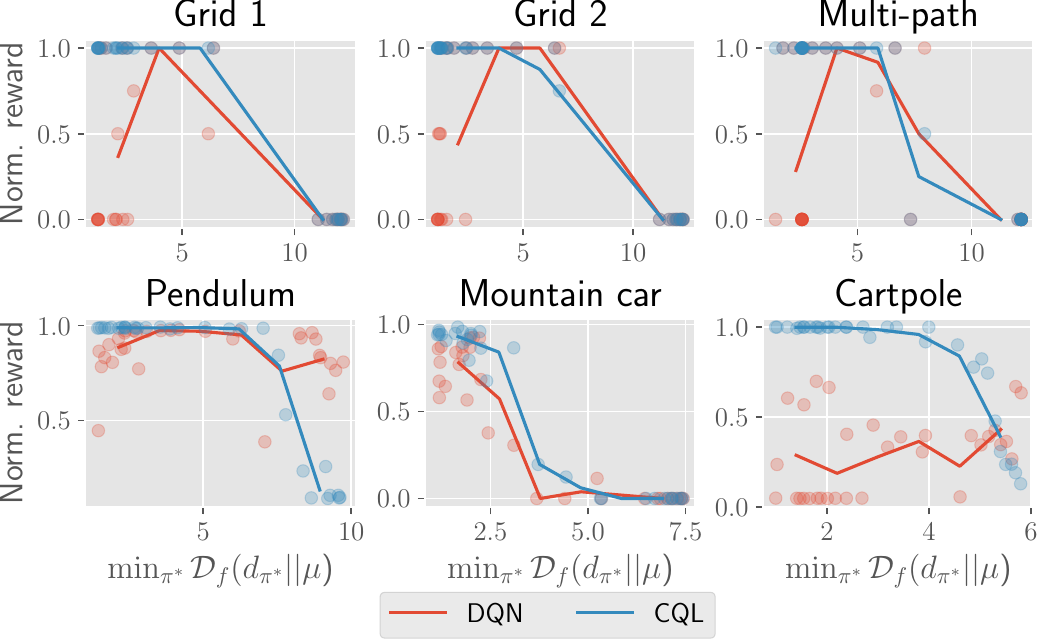}
        \caption{Dataset coverage not enforced.}
    \end{subfigure}
    \hfill
    \begin{subfigure}[b]{0.49\textwidth}
        \centering
        \includegraphics[width=0.99\textwidth]{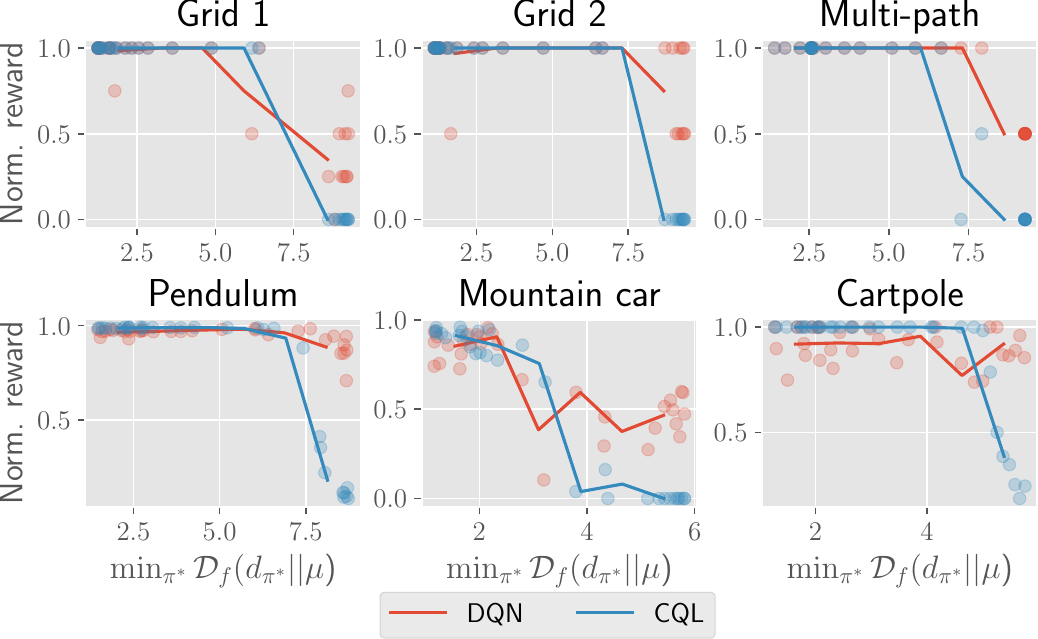}
        \caption{Dataset coverage enforced.}
    \end{subfigure}
    \caption{Average normalized rollouts reward; the x-axis encodes the statistical distance between $\mu$ and the closest distribution induced by one of the optimal policies, $d_{\pi^*}$.}
    \label{fig:exp_results:distance_plot}
\end{figure*}

The obtained experimental results are portrayed in Fig. \ref{fig:exp_results:distance_plot}, which shows the average normalized rollouts reward for different distances between $\mu$ and the closest distribution induced by one of the optimal policies, $d_{\pi^*}$. We consider a wide spectrum of behavior policies, from optimal to anti-optimal policies (i.e., Boltzmann policies with low $T$ values), as well as from fully exploitatory to fully exploratory policies. As can be seen, as the statistical distance between distribution $\mu$ and the closest distribution induced by one of the optimal policies increases, the lower the rewards obtained, irrespectively of the algorithm. We can also observe an increase in obtained rewards when dataset coverage is enforced (Fig. \ref{fig:exp_results:distance_plot} (b)) in comparison to when dataset coverage is not enforced (Fig. \ref{fig:exp_results:distance_plot} (a)).

At first sight, our results appear intuitive if we focus on Fig. \ref{fig:exp_results:distance_plot} (a), where dataset coverage is not enforced: if the policy used to collect the dataset is not good enough, it will fail to collect trajectories rich in rewards, key to learn reward-maximizing behavior. As an example, if the policy used to collect the data is highly exploratory, the agent will likely not reach high rewarding states and the learning signal may be too weak to learn an optimal policy.

However, the results displayed in Fig. \ref{fig:exp_results:distance_plot} (b), in which dataset coverage is enforced, reveal a rather less intuitive finding: despite the fact that all datasets feature full coverage over $\mathcal{S} \times \mathcal{A}$, if the statistical distance between the two distributions is high, we observe a deterioration in algorithmic performance. In other words, despite the fact that the datasets contain all the information that can be retrieved from the environment (including transitions rich in reward), offline learning can still struggle if the behavior policy is too distant from the optimal policy. Such observation can be explained by the fact that distributions far from the optimal policy prevent the propagation of information, namely $Q$-values, during the execution of the offline RL algorithm.

Given the experimental results presented in this section, it is important for the data distribution to be aligned with that of optimal policies, not only to ensure that trajectories are rich in reward, but also to mitigate algorithmic instabilities. Our experimental results suggest that the assumption of a bounded concentrability coefficient, as discussed in Sec. \ref{sec:the_data_distribution_matters:theoretical_bounds}, may not be enough to robustly learn in an offline manner and that more stringent assumptions on the data distribution are required. \citet{wang_2020} reach a similar conclusion, from a theoretical perspective.

\subsection{Discussion}
This section experimentally assessed the impact of different data distribution properties in the performance of offline $Q$-learning algorithms with function approximation, showing that the data distribution greatly impacts algorithmic performance. In summary, our results show that: (i) high entropy data distributions are well-suited for learning in an offline manner; (ii) a certain degree of data diversity (data coverage) is desirable for offline learning; and (iii) a certain degree of data quality (closeness to distributions induced by optimal policies) is desirable for offline learning.

Finding (i) is aligned with the discussion in Sec.~\ref{sec:the_data_distribution_matters:theoretical_bounds}: in the absence of detailed information regarding the underlying MDP, high entropy distributions contribute to high coverage over the state-action space, thus yielding bounded concentrability coefficients (an assumption widely adopted by the works surveyed in Sec.~\ref{sec:the_data_distribution_matters:theoretical_bounds}). However, as our experiments in Sec.~\ref{sec:experimental_results:closeness_to_opt_policy} show, full coverage (equivalent to having bounded concentrability coefficients) is not enough to learn optimal policies. Thus, we hypothesize that the advantages of using high entropy distributions not only come from the fact that they yield high coverage over the state-action space, but also because they induce smooth distributions that mitigate information bottlenecks during algorithm execution, allowing $Q$-values to easily propagate according to the MDP dynamics. This hypothesis is supported by the fact that, as we show in Proposition \ref{theo:minimax}, the maximum entropy distribution is the one that minimizes the statistical distance to all other possible distributions.

Regarding finding (ii), a certain degree of coverage is necessary, even when the data is collected by an optimal policy, due to bootstrapping, as already discussed in Sec.~\ref{sec:experimental_results:coverage}. However, it is interesting to note that, according to our results (Table \ref{tab:q_vals_and_rewards}), it is not necessary to have full coverage over the state-action space to learn optimal behavior. This finding is supported by the discussion in Sec.~\ref{sec:the_data_distribution_matters:theoretical_bounds}: as suggested by Eqs.~\eqref{eq:C2_bound} and \eqref{eq:supremum_coef}, the states that have positive probability according to the distribution induced by the optimal policy are those we care the most about; all other states are exponentially less important depending on their distance to the states belonging to the optimal trajectory (in terms of the number of MDP transitions). This contrasts with the theoretical works surveyed in Sec.~\ref{sec:the_data_distribution_matters:theoretical_bounds} that, in the absence of knowledge regarding the exact underlying MDP and the quality of the offline trajectories provided, assume bounded concentrability coefficients, i.e., full coverage over the state-action space.

\begin{table}[t]
\begin{center}
%\begin{minipage}%{174pt}
\caption{Performance metrics for the \textit{grid 1} and \textit{mountain car} environments under the DQN algorithm (coverage not enforced). For reference, the maximum average $Q$-values error recorded across all tested dataset types under the \textit{grid 1} and \textit{mountain car} environments are, respectively, $2.45 \times 10^5$ and $50.95$.}\label{tab:q_vals_and_rewards}%
\begin{tabular}{@{}lllll@{}}
\toprule
\textbf{Environment} & \textbf{Dataset} & \textbf{Avg. dataset} & \textbf{Norm. avg.} & \textbf{Avg. $Q$-values} \\
 &  \textbf{type} & \textbf{coverage} & \textbf{reward} & \textbf{error} \\
\toprule
\textit{Grid 1}   & \textit{Uniform}  & 1.0 & 1.0  & 0.05 \\
   & \textit{Boltzmann}($4.0$)  & 0.92 & 1.0  & 0.04 \\
         & ($\epsilon=0.6$)-\textit{greedy} & 0.87 & 1.0  & 0.14  \\ \midrule
\textit{Mountain car} & \textit{Uniform}  & 1.0 & 0.94 & 1.98 \\
 & ($\epsilon=0.3$)-\textit{greedy} & 0.68 & 0.97 & 2.45  \\
         & \textit{Boltzmann}($2.0$) & 0.61 & 0.91 & 3.49 \\
\botrule
\end{tabular}
%\end{minipage}
\end{center}
\end{table}

Finally, according to finding (iii), it appears to also be important that the distribution induced by the dataset is not very far from the distribution induced by one of the optimal policies of the MDP, even if all state-action pairs are present in the dataset. Again, we hypothesize that this finding can be explained by the fact that certain distributions prevent the propagation of $Q$-values throughout the iterations of the algorithm. Further investigations should be carried out in order to understand if this problem can be circumvented by using more sophisticated sampling techniques such as prioritized replaying. We leave such research direction for future work.

\paragraph{On the optimism vs pessimism tradeoff}
As seen in Fig. \ref{fig:exp_results:distance_plot}, the performance of DQN and CQL is dependent on whether coverage is enforced or not. When dataset coverage is not enforced (Fig. \ref{fig:exp_results:distance_plot} (a)), CQL outperforms DQN, especially for distributions close to that of optimal policies (as can also be seen in Fig. \ref{fig:exp_results:coverage} (a)). However, when coverage is enforced (Fig. \ref{fig:exp_results:distance_plot} (b)), DQN outperforms CQL, especially for distributions that are more distant to that of optimal policies, as can be seen in Fig. \ref{fig:exp_results:distance_plot} (b). This is due to the fact that both algorithms balance the tradeoff between optimism and pessimism in different ways. DQN is very optimistic and fails under low-coverage settings, since it propagates erroneous $Q$-values during the execution of the algorithm. However, due to its optimistic nature, it outperforms CQL when coverage is enforced, taking advantage of information that is underrepresented in the dataset. CQL, on the other hand, outperforms DQN under low coverage settings since, due to its pessimistic nature, prevents the propagation of erroneous $Q$-values. However, when valuable but underrepresented information is present in the dataset, the pessimism of CQL prevents learning, and CQL is outperformed by DQN.

\paragraph{On the impact of the sampling error, approximation capacity, and generalization hardness}
In our experiments, we consider environments featuring low sampling error, i.e., the environments have almost deterministic transitions. We also consider large-capacity function approximators, i.e., the approximators have enough capacity to represent the optimal $Q$-function exactly. Naturally, we expect our results to change if these assumptions were to change. Namely, more samples per state-action pair would be required as the stochasticity of the environment increases. If the capacity of the approximator decreases, we expect the approximator to need to focus on a subset of state-action pairs that are more relevant to optimal behavior rather than correctly estimating the $Q$-values for all state-action pairs. Finally, in order to better study the impact of different data distributions properties in the performance of ADP-related algorithms, we considered three environments, namely the \textit{grid 1}, \textit{grid 2}, and \textit{multi-path} environments, which feature highly uncorrelated features. We expect the performance of ADP-related algorithms to be less affected by changes to the data distribution under rather smoother features. Nevertheless, our main findings appear to be consistent for the remainder environments, namely the \textit{pendulum}, \textit{cartpole} and \textit{mountaincar} environments, which comprise features that allow to more easily generalize across the state-action space.

\section{Conclusion}
\label{sec:conclusion}
In this work, we investigate the interplay between the data distribution and $Q$-learning-based algorithms with function approximation. We analyze how different properties of the data distribution affect performance in both online and offline RL settings. We show, both theoretically and empirically, that: (i) high entropy data distributions contribute to mitigate sources of algorithmic instability; and (ii) different properties of the data distribution influence the performance of RL methods with function approximation. We provide a thorough experimental assessment of the performance of both DQN and CQL algorithms under several types of offline datasets, connecting our experimental results with the theoretical findings of previous works.

The experimental results presented herein provide useful insights for the development of improved data processing techniques for offline RL, which should be valuable for future research. For example, our results suggest that maximum entropy exploration methods \citep{hazan_2018} can be well suited for the construction of datasets for offline RL, naive dataset concatenation can lead to deterioration in performance, and that, by simply reweighting or discarding of training data, it is possible to substantially improve performance of offline RL algorithms.

\backmatter

%\bmhead{Supplementary information}

%If your article has accompanying supplementary file/s please state so here. 

%Authors reporting data from electrophoretic gels and blots should supply the full unprocessed scans for key as part of their Supplementary information. This may be requested by the editorial team/s if it is missing.

%Please refer to Journal-level guidance for any specific requirements.

\section*{Acknowledgments}
This work was partially supported by Portuguese national funds through the Portuguese Fundação para a Ciência e a Tecnologia (FCT) under projects UIDB/50021/2020 (INESC-ID multi-annual funding), PTDC/CCI-COM/5060/2021 (RELEvaNT), and PTDC/CCI-COM/7203/2020 (HOTSPOT). In addition, this research was partially supported by TAILOR, a project funded by EU Horizon 2020 research and innovation programme under GA No. 952215, and by the Air Force Office of Scientific Research under award number FA9550-22-1-0475. Pedro P. Santos acknowledges the FCT PhD grant 2021.04684.BD. Diogo S. Carvalho acknowledges the FCT PhD grant 2020.05360.BD.

%\subsection*{Competing interests}
%
%The authors have no relevant financial or non-financial interests to disclose.

%\subsection*{Ethics Approval}
%
%Not applicable.

%\subsection*{Consent to participate}
%
%Not applicable.

%\subsection*{Consent for publication}
%
%Not applicable.

%\subsection*{Availability of data and material}
%
%All experimental data is publicly available at \url{https://github.com/PPSantos/rl-data-distribution-public}. A dashboard containing the experimental results can be accessed at \url{https://rldatadistribution.pythonanywhere.com/}

%\subsection*{Code availability}
%
%The developed code is publicly available at \url{https://github.com/PPSantos/rl-data-distribution-public}

%\subsection*{Authors' contributions}
%
%The authors, Pedro P. Santos, Diogo S. Carvalho, Alberto Sardinha, and Francisco S. Melo, have all contributed to all parts of the research (theory, experiments, and writing).

%%===========================================================================================%%
%% If you are submitting to one of the Nature Portfolio journals, using the eJP submission   %%
%% system, please include the references within the manuscript file itself. You may do this  %%
%% by copying the reference list from your .bbl file, paste it into the main manuscript .tex %%
%% file, and delete the associated \verb+\bibliography+ commands.                            %%
%%===========================================================================================%%

\bibliography{biblio}% common bib file
%% if required, the content of .bbl file can be included here once bbl is generated
%%\input sn-article.bbl

%% Default %%
%%\input sn-sample-bib.tex%

%\section*{Declarations}

%Some journals require declarations to be submitted in a standardised format. Please check the Instructions for Authors of the journal to which you are submitting to see if you need to complete this section. If yes, your manuscript must contain the following sections under the heading `Declarations':

%\begin{itemize}
%\item Funding
%\item Conflict of interest/Competing interests (check journal-specific guidelines for which heading to use)
%\item Ethics approval 
%\item Consent to participate
%\item Consent for publication
%\item Availability of data and materials
%\item Code availability 
%\item Authors' contributions
%\end{itemize}

%\noindent
%If any of the sections are not relevant to your manuscript, please include the heading and write `Not applicable' for that section. 

%%===================================================%%
%% For presentation purpose, we have included        %%
%% \bigskip command. please ignore this.             %%
%%===================================================%%
%\bigskip
%\begin{flushleft}%
%Editorial Policies for:

%\bigskip\noindent
%Springer journals and proceedings: \url{https://www.springer.com/gp/editorial-policies}

%\bigskip\noindent
%Nature Portfolio journals: \url{https://www.nature.com/nature-research/editorial-policies}

%\bigskip\noindent
%\textit{Scientific Reports}: \url{https://www.nature.com/srep/journal-policies/editorial-policies}

%\bigskip\noindent
%BMC journals: \url{https://www.biomedcentral.com/getpublished/editorial-policies}
%\end{flushleft}

\newpage
\begin{appendices}

\section{Supplementary Materials for Section \ref{sec:background}}
\label{sec:supplementarysec2}
Below, we present the pseudocode of a generic $Q$-learning algorithm with function approximation.
\begin{algorithm}
\caption{Generic $Q$-learning algorithm with function approximation.}\label{algo:DQN}
\begin{algorithmic}[1]
\State \text{initialize} $\phi_0$
\State \text{initialize} $s_0 \sim p_0$
\State \text{initialize} $\mathcal{D} = \emptyset$%\Comment{\text{Initialize replay memory.}}
\For{\text{iteration} $k$ in $\{K\}$}
\For{\text{sampling step in} $\{S\}$}
\State $a_t \sim \pi_k(a_t \rvert s_t)$
\State $s_{t+1} \sim p(s_{t+1} \rvert s_{t},a_{t})$
\State $\mathcal{D} = \mathcal{D} \cup \{(s_t,a_t,r(s_t,a_t),s_{t+1})\}$
\EndFor
\State $\phi_{k,0} = \phi_k$
\For{\text{gradient step} $g$ in $\{G\}$}
\State \text{sample batch $\{(s^i_t,a^i_t,r^i_t,s^i_{t+1})\}$ from $\mathcal{D}$}
\State $\phi_{k,g+1} \leftarrow \phi_{k,g} - \alpha \nabla_{\phi_{k,g}} \sum_i$
\State \quad \quad $\left( Q_{\phi_{k,g}}(s^i_t,a^i_t) - (r^i_t + \gamma \max_{a'}Q_{\phi_k}(s^i_{t+1}, a'))\right)^2$
\EndFor
\State $\phi_{k+1} = \phi_{k,G}$
\EndFor
\end{algorithmic}
\end{algorithm}

\section{Supplementary Materials for Section \ref{sec:the_data_distribution_matters}}
\label{sec:supplementarysec4}

This section gives additional details with respect to Section \ref{sec:the_data_distribution_matters}.

\subsection{Supplementary Materials for Section \ref{sec:the_data_distribution_matters:theoretical_bounds}}
As noted in Section \ref{sec:the_data_distribution_matters:theoretical_bounds}, concentrability coefficient $C_3$, as defined by \eqref{eq:C2_bound} and \eqref{eq:expectation_coef}, is equivalent to an $f$-divergence between a given distribution $\beta$ and the sampling distribution $\mu$. As stated, that is the case when $f(x) = x^2 -1$, since
\begin{align*}
 \left\| \frac{\beta}{\mu}\right\|_{2,\mu} &= \left(\mathbb{E}_{(s,a)\sim \mu}\left[\left(\frac{\beta(s,a)}{\mu(s,a)}\right)^2\right]\right)^{1/2} \\
 &= \left(\mathbb{E}_{(s,a)\sim \mu}\left[\left(\frac{\beta(s,a)}{\mu(s,a)}\right)^2\right] -1 + 1\right)^{1/2} \\
 &= \left(\mathbb{E}_{(s,a)\sim \mu}\left[\left(\frac{\beta(s,a)}{\mu(s,a)}\right)^2 - 1\right] + 1\right)^{1/2} \\
 &= \left(\mathbb{E}_{(s,a)\sim \mu}\left[f\left(\frac{\beta(s,a)}{\mu(s,a)}\right)\right] + 1\right)^{1/2} \\
 &=\sqrt{\mathcal{D}_{f}(\beta \rvert\rvert \mu) + 1}\\
 &=\sqrt{\chi^2(\beta \rvert\rvert \mu) + 1}.
\end{align*}
As the last equality states, the $f$-divergence above corresponds to a particular type of divergence known as the $\chi^2$-divergence \citep{liese_2006}.\\

Below, we prove Proposition \ref{theo:minimax}. We reproduce the proposition's statement before the proof for ease of consultation.

\begin{proposition}
Let $\mathcal{P}(\mathcal{S}, \mathcal{A})$ represent the set of probability distributions over $\mathcal{S} \times \mathcal{A}$.%, also known as the probability simplex. 
Let also $L_\mu: \mathcal{P}(\mathcal{S} \times \mathcal{A}) \to \mathbb{R}^+_0$ such that $L_\mu(\beta) = \left\| \frac{\beta}{\mu} \right\|_{2, \mu}^2$.
The solution $\mu$ to %the minimax objective
\begin{equation}%\label{game:minimax}
    \argmin_{\mu \in \mathcal{P}(\mathcal{S} \times \mathcal{A})} \max_{\beta \in \mathcal{P}(\mathcal{S} \times \mathcal{A})} L_\mu(\beta)
\end{equation}
is the uniform distribution over the state-action space.%, $\mathcal{U}(\mathcal{S}\times\mathcal{A})$.
\end{proposition}

\begin{proof}
Suppose that, as discussed, the maximizing player (adversary) has access to a distribution $\mu$, chosen by the minimizing player. The adversary's objective is therefore to maximize $L_\mu$ over the probability simplex. We begin to show what is the solution of such maximization. We start by noting that $L_\mu$, being a norm, is a convex real function. $L_\mu$ is also continuous. Additionally, the probability simplex $\mathcal{P}(\mathcal{S}\times\mathcal{A})$ is a compact set, since it is both closed and bounded. Under the three conditions just mentioned,
%, on the objective function $L_\mu$ and the admissible region $\mathcal{P}(\mathcal{S}\times\mathcal{A}), $ 
Bauer's Maximum Principle guarantees that the solution of the maximization lies on the subset of extreme points of the admissible region. Such set is, in the case of the probability simplex, the subset of probability distributions with a singleton support set. Equivalently, the adversary is to choose a distribution $\beta$ such that, for some pair $(s, a)$, $\beta(s, a) = 1$ and is zero otherwise. Formally, we can write
\begin{align} 
 L_\mu(\beta) &= \sum_{(s, a) \in \mathcal{S} \times \mathcal{A}} \mu(s, a) \left(\frac{\beta(s, a)}{\mu(s, a)}\right)^2 \\
&= \sum_{(s, a) \in \mathcal{S} \times \mathcal{A}} \mu(s, a)^{-1} \beta(s, a)^2.
\end{align} 
Therefore,
\begin{align}
    \max_{\beta \in \mathcal{P}(\mathcal{S} \times \mathcal{A})} L_\mu(\beta) &= \max_{(s, a)  \in \mathcal{S} \times \mathcal{A}} \mu(s, a)^{-1}. 
\end{align}
Finally, the minimizing player's best choice of $\mu$ is the one minimizing the quantity above. We note that
\begin{equation}
\max_{(s, a) \in \mathcal{S} \times \mathcal{A}} \mu(s, a)^{-1} = \lvert \mathcal{S} \rvert \times \lvert \mathcal{A} \rvert
\end{equation} 
if $\mu$ is the uniform distribution and
\begin{equation}
\max_{(s, a) \in \mathcal{S} \times \mathcal{A}} \mu(s, a)^{-1} > \lvert \mathcal{S} \rvert \times \lvert \mathcal{A} \rvert
\end{equation}
otherwise. The conclusion follows.
\end{proof}

\begin{figure}[h]
\begin{center}
    \begin{tabular}{cc}
    \includegraphics[width=0.45\textwidth]{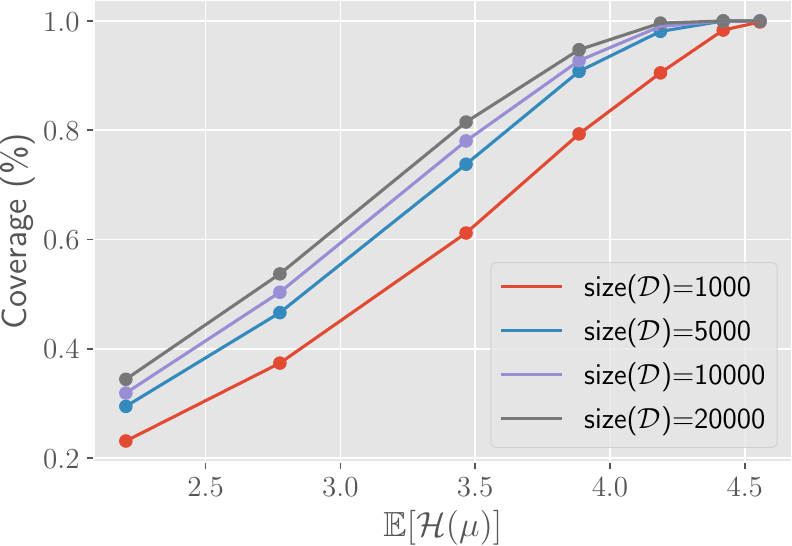} &
    \includegraphics[width=0.45\textwidth]{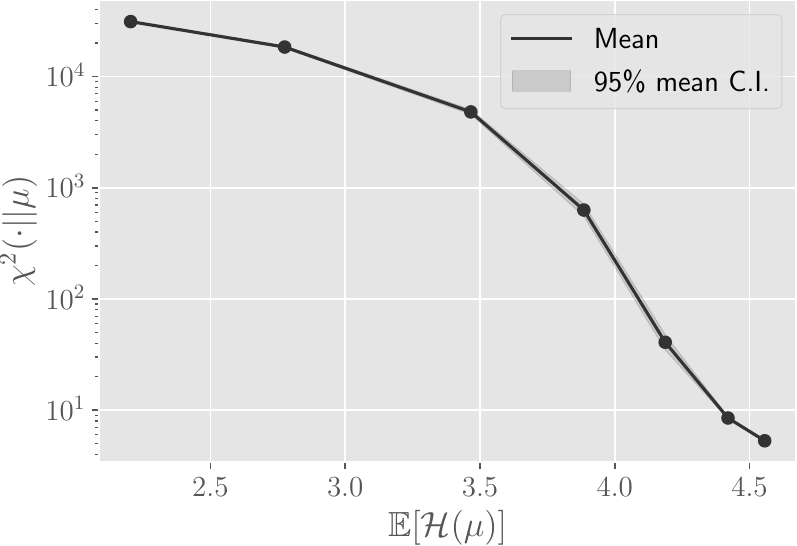} \\
    (a) Dataset coverage. & (b) Distance to all other distributions. \\
    \end{tabular}
\caption{The relationship between the expected entropy of the distribution $\mu$ and: (i) the dataset coverage; and (ii) the $\chi^2$-divergence to all other distributions.}
\label{fig:appendix:high_entropy_dists_motivation}
\end{center}
\end{figure}

Figure \ref{fig:appendix:high_entropy_dists_motivation} displays the relationship between the expected entropy of the sampling distribution $\mu$ and: (i) the coverage of different datasets constructed using $\mu$; and (ii) the mean $\chi^2$-divergence to all other distributions. The plots are computed using randomly sampled distributions, which we sample from different Dirichlet distributions. We control the expected entropy of the resulting distributions by varying the Dirichlet parameter $\alpha$. As seen in Fig. \ref{fig:appendix:high_entropy_dists_motivation} (a), irrespectively of the dataset size, higher entropy distributions lead to datasets featuring higher coverage over the distribution support. As seen in Fig. \ref{fig:appendix:high_entropy_dists_motivation} (b), higher entropy distributions yield, on average, a lower $\chi^2$-divergence to all other sampled distributions in comparison to lower entropy distributions.

\subsection{Supplementary Materials for Section \ref{sec:the_data_distribution_matters:four_state_MDP}}
\label{sec:supplementary_four_state_MDP}

\begin{figure}[h]
\centering
\includegraphics[width=0.5\textwidth]{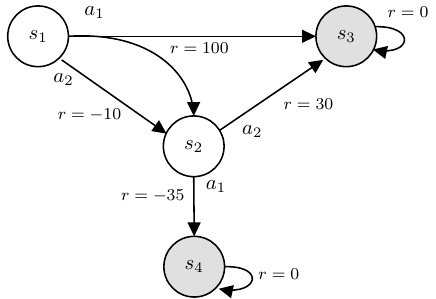}
\caption{Four-state MDP, comprising states $\{s_1,s_2,s_3,s_4\}$ and actions $\{a_1,a_2\}$. State $s_1$ is the initial state and states $s_3$ and $s_4$ are the terminal absorbing states. All actions lead to deterministic transitions, except for the state-action pair $(s_1,a_1)$, where $p(s_3 \rvert s_1,a_1) = 0.99$ and $p(s_2 \rvert s_1,a_1) = 0.01$. The reward function is defined as $r(s_1,a_1) = 100$, $r(s_1,a_2) = -10$, $r(s_2,a_1) = -35$ and $r(s_2,a_2) = 30$.}
\label{fig:four_states_mdp_appendix}
\end{figure}

In this section we study how the data distribution can influence the performance of $Q$-learning based RL algorithms with function approximation, under a four-state MDP (Fig. \ref{fig:four_states_mdp_appendix}). The main objective of this section is to show that the data distribution plays an active role regulating algorithmic stability. We show that the data distribution can significantly influence the quality of the resulting policies and affect the stability of the learning algorithm. We consider both online and offline RL settings. We finish this section by summarizing the key insights and providing a discussion on how the example presented here, as well as the respective findings, can generalize into bigger and more realistic MDPs.

All experimental results are averaged over 6 independent runs.

We focus our attention on non-terminal states $s_1$ and $s_2$ of the MDP, whose optimal $Q$-function is
\begin{equation*}
Q^*=
\begin{blockarray}{ccc}
& a_1 & a_2  \\
\begin{block}{c[cc]}
   s_1 & 100.3 & 20 \\
  s_2 & -35 & 30 \\
\end{block}
\end{blockarray}.
\end{equation*}
\noindent At state $s_1$ the correct action is $a_1$, whereas at state $s_2$ the correct action is $a_2$. We set $\gamma = 1$.

We consider a linear function approximator $Q_w(s_t,a_t) = w^\mathrm{T} \phi(s_t,a_t)$, where $w = [w_1, w_2, w_3]^\mathrm{T}$ denotes a weight column vector and $\phi$ is a feature mapping, defined as ($\phi(s_1,a_1) = [1,0,0]^\mathrm{T}$, $\phi(s_1,a_2) = [0,1,0]^\mathrm{T}$, $\phi(s_2,a_1) = [\alpha,0,0]^\mathrm{T}$, and $\phi(s_2,a_2) = [0,0,1]^\mathrm{T}$)
\begin{equation*}
Q_w=
\begin{blockarray}{ccc}
& a_1 & a_2  \\
\begin{block}{c[cc]}
   s_1 & w_1 & w_2 \\
  s_2 & \alpha w_1 & w_3 \\
\end{block}
\end{blockarray},
\end{equation*}
\noindent where $\alpha \in [1, 3/2)$. The optimal policy is retrieved if $w_1 > w_2$ and $\alpha w_1 < w_3$. As can be seen, due to the choice of feature mapping, the capacity of the function approximator is limited and there exists a correlation in the features between $Q_w(s_1,a_1)$ and $Q_w(s_2,a_1)$. This will be key to the results that follow.

\subsubsection{Offline Oracle Version}
\begin{figure}[h]
\begin{center}
    \begin{tabular}{cc}
    \includegraphics[width=0.45\textwidth]{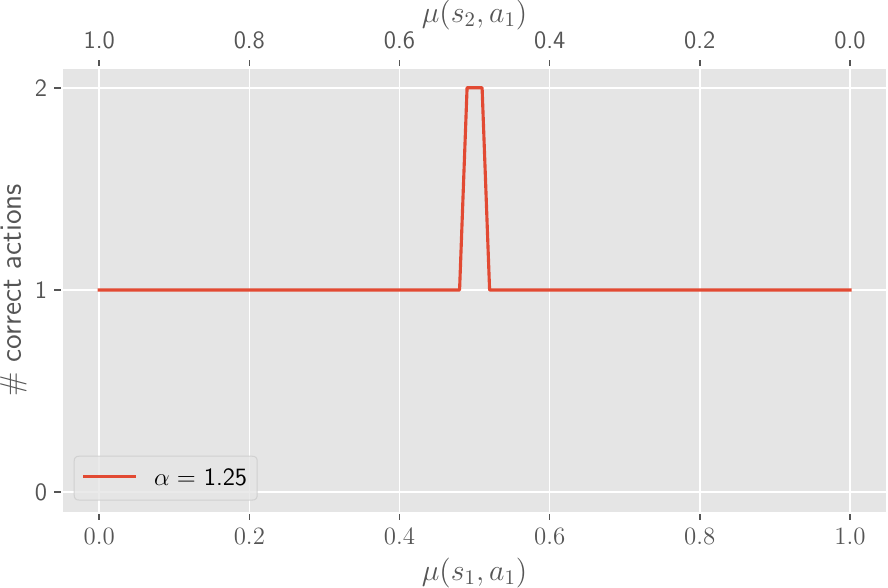} &
    \includegraphics[width=0.45\textwidth]{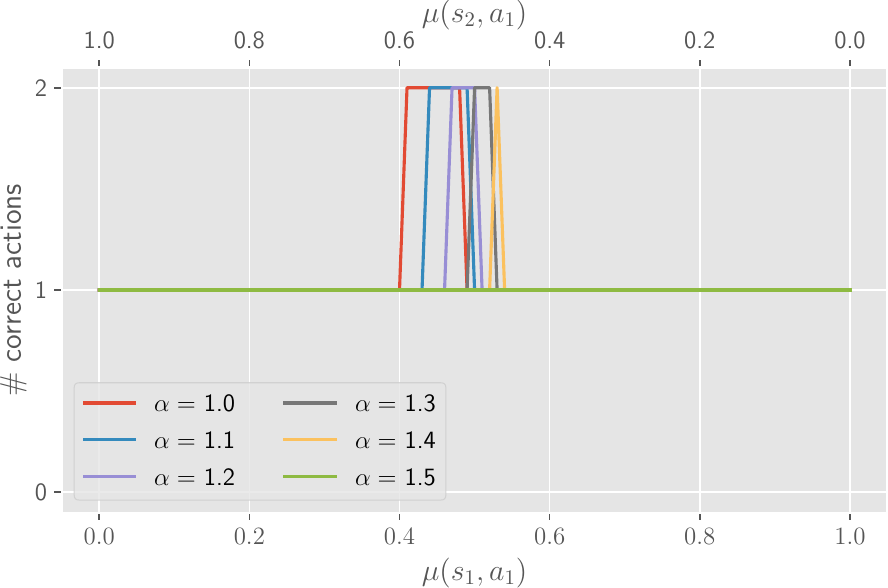} \\
    (a) $\alpha = \{ 1.25 \}$. & (b) $\alpha = \{ 1.0, 1.1, 1.2, 1.3, 1.4, 1.5 \}$. \\
    \end{tabular}
\caption{The number of correct actions at states $s_1$ and $s_2$ of the four-state MDP for different data distributions. The number of correct actions is calculated using Eqs. \ref{eq:oracle_optimal_weights} for different $\mu(s_1,a_1)$ and $\mu(s_1,a_2)$ proportions.}
\label{fig:correct_actions}
\end{center}
\end{figure}

We start by assuming that we have access to an oracle providing us with the exact optimal $Q$-function and write the loss of the function approximator as
\begin{equation*}
\mathcal{L}(w) = \mathbb{E}_{(s_t,a_t)\sim \mu}\left[ \left(w^\mathrm{T} \phi(s_t,a_t) - Q^*(s_t,a_t) \right)^2 \right],
\end{equation*}
\noindent where $\mu$ can be interpreted as the probability distribution over state-action pairs induced by a static dataset of transitions or a generative model of the environment. The optimal weight vector is given by
\begin{align}
  &\begin{aligned}
  \label{eq:oracle_optimal_weights}
  w_1^* &= \frac{\mu(s_1,a_1)Q^*(s_1,a_1) + \alpha \mu(s_2,a_1)Q^*(s_2,a_1)}{\mu(s_1,a_1)+\alpha^2\mu(s_2,a_1)}, \\
  w_2^* &= Q^*(s_1,a_2), \\
  w_3^* &= Q^*(s_2,a_2),
  \end{aligned}
\end{align}
\noindent as long as $\mu(s,a) > 0, \forall (s,a)$. In the results that follow, we focus our attention on the proportion between probabilites $\mu(s_2,a_1)$ and $\mu(s_1,a_1)$, thus setting $\mu(s_2,a_1) = 1 - \mu(s_1,a_1)$. Figure \ref{fig:correct_actions} (a) displays the influence of the data distribution, namely the proportion between $\mu(s_1,a_1)$ and $\mu(s_2,a_1)$, on the learned policy when $\alpha=5/4$. As can be seen, for the present MDP, the optimal policy is only attained if  $\mu(s_1,a_1) \approx 0.5$ and $\mu(s_2,a_1) \approx 0.5$. Precisely, as aforementioned, the optimal policy is retrieved if $w_1^* > w_2^*$ and $\alpha w_1^* < w_3^*$. Such inequalities are verified when
\begin{align*}
  &\begin{aligned}
  \frac{\alpha(20\alpha+35)}{80.3+35\alpha+20\alpha^2} &< \mu(s_1,a_1) < \frac{65\alpha^2}{65\alpha^2+100.3\alpha-30}\\
  (\approx 0.48) &< \mu(s_1,a_1) < (\approx 0.51).
  \end{aligned}
\end{align*}
\noindent However, as can be seen in the figure, if $\mu(s_1,a_1)$ is not $\approx 0.5$, the optimal policy will not be retrieved. As an example, if $\mu(s_1,a_1)= 0.7$ and $\mu(s_1,a_2) = 0.3$, we have $w_1^* \approx 48.8$, $w_2^* = 20$ and $w_3^* = 30$. Thus, the policy is correct at state $s_1$ since $w_1^* > w_2^*$, but wrong at state $s_2$ since condition $\alpha w_1^* < w_3^*$ is not verified. We observe a somewhat similar trend for $\alpha \in [1, 3/2) $, as can be seen in Figure \ref{fig:correct_actions} (b): the retrieval of the optimal policy is dependent on $\mu(s_1,a_1)$ and $\mu(s_2,a_1)$.

The results above show that, due to the limited approximation power and correlation between features, the data distribution impacts the performance of the resulting policies, even for the presently considered oracle-based algorithm (with access to the true $Q^*$ function). The number of retrieved correct actions is directly dependent on the properties of the data distribution.

\subsubsection{Offline Temporal Difference Version}
Naturally, practical algorithms do not have access to the unknown target, $Q^*$. A common choice of replacement is to use the one-step temporal difference target. In such cases, the loss function becomes
\begin{equation*}
\mathcal{L}(w) = \mathbb{E}_{(s_t,a_t,s_{t+1})\sim \mu}\left[ \left( r(s_t,a_t) + \max_{a' \in \mathcal{A}} w^\mathrm{T}\phi(s_{t+1},a') - w^\mathrm{T} \phi(s_t,a_t)\right)^2 \right].
\end{equation*}
If we neglect the influence of the weight vector in the target (semi-gradient), the update equation for the weight vector is given by
\begin{align*}
  &\begin{aligned}
  w_{t+1} &= w_t - \eta \frac{\partial \mathcal{L}(w)}{\partial w}\\
  &= w_t + \eta \mathbb{E} \left[ \left( r(s_t,a_t) + \max_{a' \in \mathcal{A}} w_t^\mathrm{T}\phi(s_{t+1},a') - w_t^\mathrm{T} \phi(s_t, a_t) \right) \phi(s_t,a_t)\right].\\
  \end{aligned}
\end{align*}
\noindent where $\eta$ denotes the learning rate. Component-wisely,
\begin{align}
  \label{eq:TD_weights_update}
  &\begin{aligned}
  w_{1,t+1} &= w_{1,t} + \eta \mu(s_1,a_1) \left( r(s_1,a_1) + p(s_2 \rvert a_1, a_1) \max \{Q_w(s_2,a_1), Q_w(s_2,a_2)\} - w_{1,t} \right)\\
    &+ \eta \mu(s_2,a_1) \left( r(s_2,a_1) - \alpha w_{1,t} \right)\alpha \\
  &= w_{1,t} + \eta \mu(s_1,a_1) \left( r(s_1,a_1) + p(s_2 \rvert a_1, a_1) \max \{\alpha w_{1,t}, w_{3,t}\} - w_{1,t} \right)\\
    &+ \eta \mu(s_2,a_1) \left( r(s_2,a_1) - \alpha w_{1,t} \right)\alpha, \\
  w_{2,t+1} &= w_{2,t} + \eta \mu(s_1,a_2) \left( r(s_1,a_2) + \max\{Q_w(s_2,a_1), Q_w(s_2,a_2)\} - w_{2,t} \right) \\
  &\propto w_{2,t} + \eta \left( r(s_1,a_2) + \max\{\alpha w_{1,t}, w_{3,t}\} - w_{2,t} \right), \\
  w_{3,t+1} &= w_{3,t} + \eta \mu(s_2,a_2) \left(r(s_2,a_2) -  w_{3,t}\right)\\
  &\propto w_{3,t} + \eta \left(r(s_2,a_2) - w_{3,t}\right).
  \end{aligned}
\end{align}

\begin{figure}[h]
\begin{center}
    \begin{tabular}{cc}
    \includegraphics[width=0.47\textwidth]{correct_actions_TD_target.pdf} &
    \includegraphics[width=0.47\textwidth]{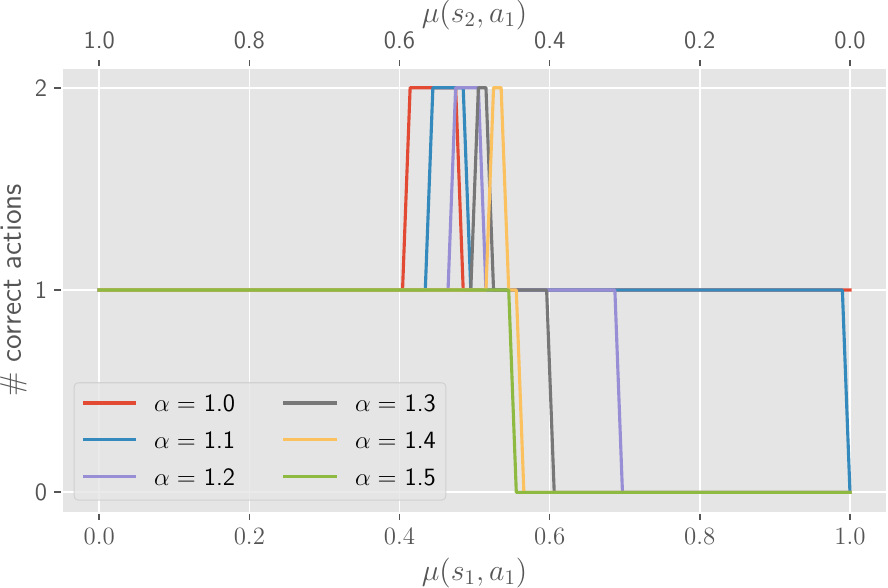} \\
    (a) $\alpha = \{ 1.25 \}$. & (b) $\alpha = \{ 1.0, 1.1, 1.2, 1.3, 1.4, 1.5 \}$. \\
    \end{tabular}
\caption{The number of correct actions at states $s_1$ and $s_2$ of the four-state MDP for different data distributions. The number of correct actions is calculated by finding the fixed-point of Eqs. \ref{eq:TD_weights_update} for different $\mu(s_1,a_1)$ and $\mu(s_1,a_2)$ proportions.}
\label{fig:correct_actions_TD_target}
\end{center}
\end{figure}

Figure \ref{fig:correct_actions_TD_target} (a) displays the influence of the data distribution, namely the proportion between $\mu(s_1,a_1)$ and $\mu(s_2,a_1)$, on the retrieved policy. As can be seen, we identify three regimes: (i) whenever $\mu(s_1,a_1) \approx 0.5$, we retrieve the optimal policy (i.e., the actions are correct at both states $s_1$ and $s_2$); (ii) if $\mu(s_1,a_1) < (\approx0.48)$ or $ (\approx 0.52) < \mu(s_1,a_1) < (\approx 0.65)$, the policy is only correct at one of the states; (iii) if $\mu(s_1,a_1) > (\approx 0.65)$, the policy is wrong at both states. % Similarly, as seen in Fig. \ref{fig:correct_actions_TD_target} (b), if $\alpha \in [1, 3/2)$, high entropy distributions will more likely retrieve the optimal policy.

Akin to the oracle version (Eqs. \ref{eq:oracle_optimal_weights}), the data distribution plays a key role in the performance of the resulting policies. However, for the present temporal difference target, the impact of the data distribution is enhanced due to the dependence of the target on the weight vector. As an example, if $\mu(s_1,a_1)= 0.7$ and $\mu(s_1,a_2) = 0.3$, we have $w_1^* \approx 49.0$, $w_2^* \approx 51.3$ and $w_3^* = 30$. As $\mu(s_1,a_1)$ increases, not only $w_1$ increases (similarly to the oracle version), but also $w_2$ wrongly increases. This is due to the fact that the target used in the estimation of $w_2$ depends on $w_1$ (Eqs. \ref{eq:TD_weights_update}).

\subsubsection{Online Temporal Difference Version with Unlimited Replay Capacity}
Reinforcement learning algorithms usually collect data in an online fashion, by using an exploratory policy such as an $\epsilon$-greedy policy with respect to the current $Q$-values estimates. We now focus on such setting, showing that similar problems arise for the previously presented MDP and set of features. Instead of considering a fixed $\mu$ distribution, we now consider a setting where the $\mu$ distribution is dynamically induced by a replay buffer. We focus our attention on the commonly used $\epsilon$-greedy exploration.

\begin{figure*}[t]
    \centering
    \begin{subfigure}[t]{0.4\textwidth}
        \centering
        \includegraphics[width=0.99\textwidth]{correct_actions_different_dists.pdf}
        \caption{Number of correct actions.}
    \end{subfigure}%
    ~ 
    \begin{subfigure}[t]{0.41\textwidth}
        \centering
        \includegraphics[width=0.99\textwidth]{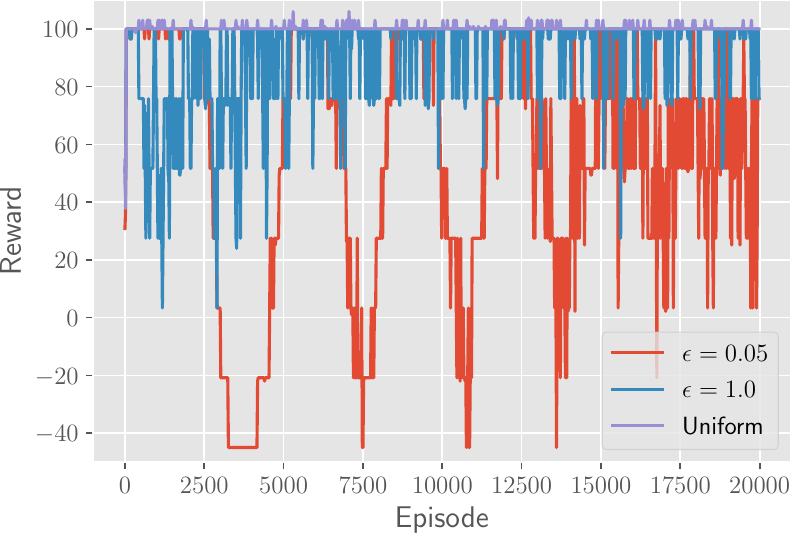}
        \caption{Greedy policy reward.}
    \end{subfigure}
    \begin{subfigure}[t]{0.4\textwidth}
        \centering
        \includegraphics[width=0.99\textwidth]{q_values_mean_error.pdf}
        \caption{$Q$-values mean error.}
    \end{subfigure}
    \caption{Four-state MDP experiments for different exploratory policies with an infinitely-sized replay buffer.}
    \label{fig:four_states_mdp_exps}
\end{figure*}

\begin{figure*}[t]
    \centering
    \begin{subfigure}[t]{0.4\textwidth}
        \centering
        \includegraphics[width=0.99\textwidth]{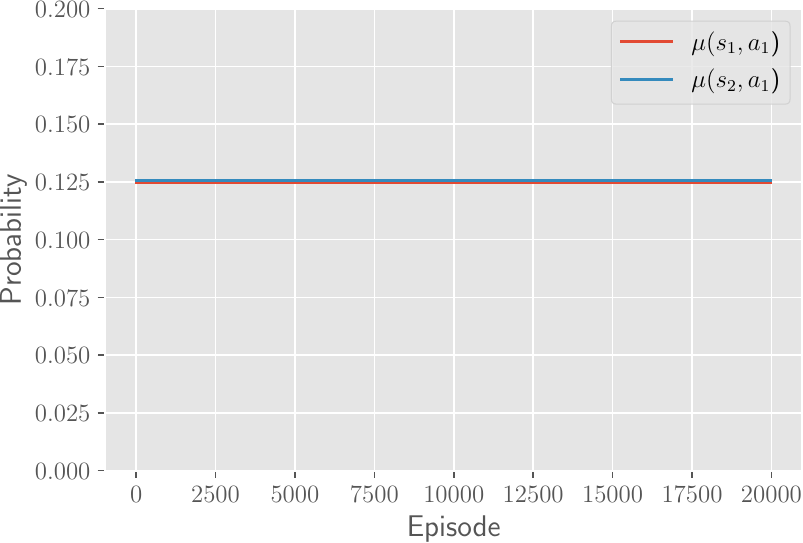}
        \caption{Uniform.}
    \end{subfigure}%
    ~ 
    \begin{subfigure}[t]{0.4\textwidth}
        \centering
        \includegraphics[width=0.99\textwidth]{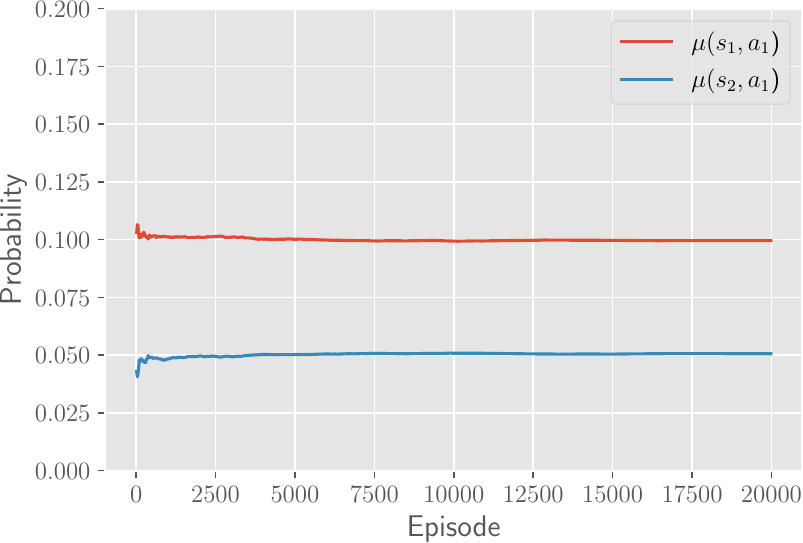}
        \caption{$\epsilon=1.0$.}
    \end{subfigure}
    \begin{subfigure}[t]{0.4\textwidth}
        \centering
        \includegraphics[width=0.99\textwidth]{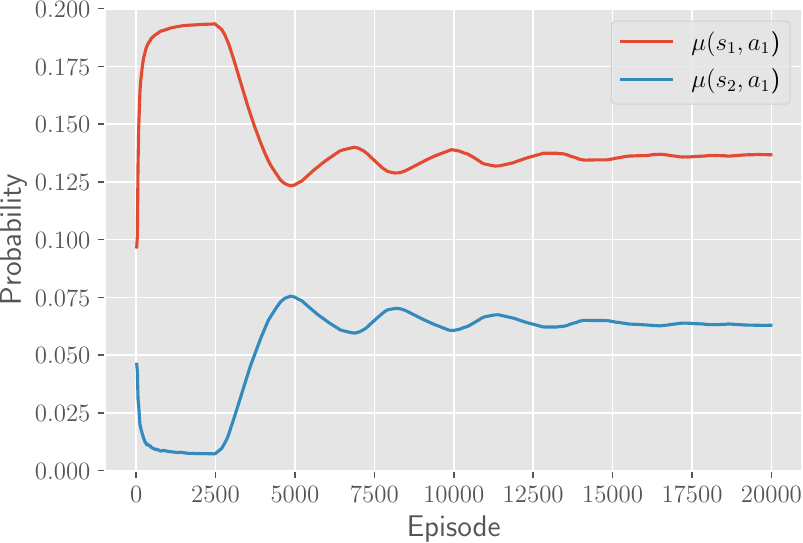}
        \caption{$\epsilon=0.05$.}
    \end{subfigure}
    \caption{Four-state MDP experiments for different exploratory policies with an infinitely-sized replay buffer. The plots display the data distribution probabilities $\mu(s_1,a_1)$ and $\mu(s_2,a_1)$, as induced by the contents of the replay buffer, throughout episodes.}
    \label{fig:four_states_mdp_exps_2}
\end{figure*}

Figures \ref{fig:four_states_mdp_exps} and \ref{fig:four_states_mdp_exps_2} display the experimental results for the four-state MDP when $\alpha = 1.2$, under two different exploratory policies: (i) a fully exploratory policy, i.e., an $\epsilon$-greedy policy with $\epsilon=1.0$; and (ii) an $\epsilon$-greedy policy with $\epsilon=0.05$. Under the current experimental setting, we consider a replay buffer size big enough such that we never discard old transitions. We use an uniform synthetic data distribution, i.e., $\mu(s_1,a_1)=\mu(s_1,a_2)=\mu(s_2,a_1)=\mu(s_2,a_2)=0.25$, as baseline.

With respect to the uniform baseline, as can be seen in Fig. \ref{fig:four_states_mdp_exps}, it outperforms all other data distributions, yielding two correct actions, higher reward, and lower $Q$-values error. This is expected, as previously discussed, since under this synthetic distribution we have that $\mu(s_1,a_1) = \mu(s_2,a_1)$.

Regarding the fully exploratory policy, i.e., the $\epsilon$-greedy policy with $\epsilon=1.0$, the agent is only able to pick the correct action at state $s_1$, featuring a lower reward and higher average $Q$-value error in comparison to the uniform distribution baseline. This is due to the fact that the stationary distribution of the MDP under the fully exploratory policy is too far from the uniform distribution to retrieve the optimal policy. As seen in Fig. \ref{fig:four_states_mdp_exps_2} (b), under the stationary distribution, we have $\mu(s_1,a_1) \approx 0.1$ and $\mu(s_2,a_1) \approx 0.05$. If we normalize the $\mu$ distribution accounting only for $\mu(s_1,a_1)$ and $\mu(s_2,a_1)$ probabilities we have that $\mu(s_1,a_1) \approx 0.7$ and $\mu(s_2,a_1) \approx 0.3$. As seen in Fig. \ref{fig:correct_actions_TD_target}, under such proportion of $\mu(s_1,a_1)$ and $\mu(s_2,a_1)$ probabilities, we only yield one correct action, as verified by the experiment.

\begin{figure}[h]
\begin{center}
    \begin{tabular}{cc}
  \includegraphics[width=0.45\textwidth]{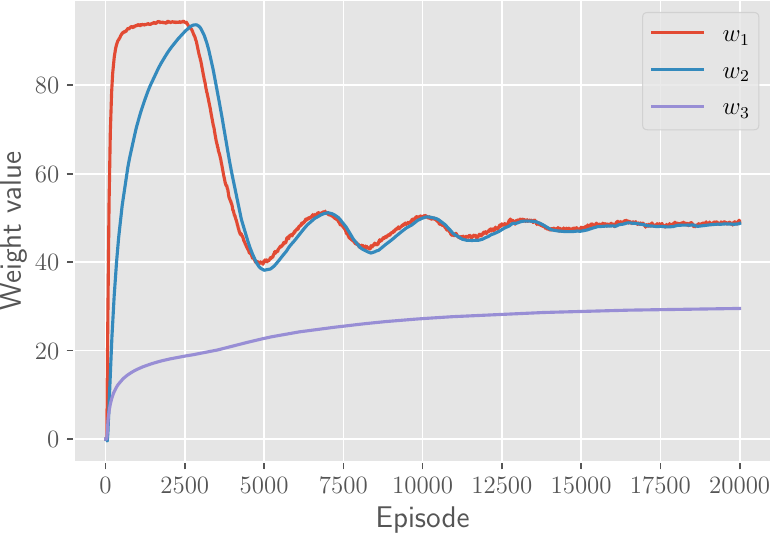} &   \includegraphics[width=0.45\textwidth]{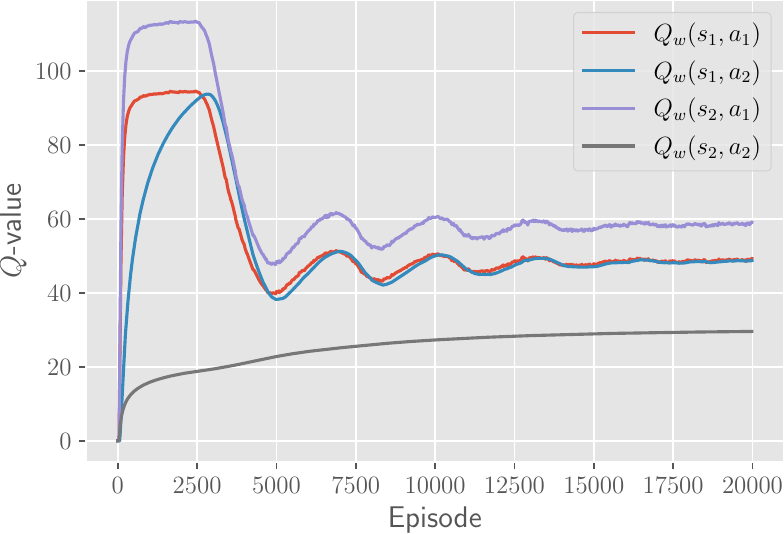}\\ 
  (a) Weights. & (b) $Q$-values. \\
\end{tabular}
\caption{Four-state MDP experiments for the $\epsilon$-greedy exploratory policy with $\epsilon=0.05$ and an infinitely-sized replay buffer. The plots display the estimated weights and $Q$-values throughout episodes.}
\label{fig:four_states_mdp_exps_q_vals}
\end{center}
\end{figure}

Finally, for the $\epsilon$-greedy policy with $\epsilon = 0.05$, the performance of the agent further deteriorates, as displayed in Fig. \ref{fig:four_states_mdp_exps}. As can be seen in Fig. \ref{fig:four_states_mdp_exps} (c) and Fig. \ref{fig:four_states_mdp_exps_q_vals}, such exploratory policy induces oscillations in the $Q$-values, which eventually damp out as learning progresses. The oscillations arise due to an undesirable interplay between the features of the function approximator and the data distribution: exploitation leads to changes in distribution $\mu$ that, in turn, drive changes in the weight vector $w$ (which implicitly regulates the action selection mechanism). The oscillations end up being dampened by the fact that the replay buffer contributes to stabilize the data distribution in such a way that exploitation will not lead to such big changes in the $\mu$ distribution. As $\mu$ approaches its stationary distribution (since the replay buffer is sufficiently large), we have that $Q_w(s_1,a_1) \approx Q_w(s_1,a_2)$ and $Q_w(s_2,a_1) > Q_w(s_2,a_2)$, as can be seen in Fig. \ref{fig:four_states_mdp_exps_q_vals} (b). We reach an undesirable solution: the agent cannot distinguish which is the best action at state $s_1$. Thus, it keeps alternating between actions $a_1$ and $a_2$ at state $s_1$, as seen in Fig. \ref{fig:four_states_mdp_exps} (a).

\paragraph{Detailed description of the interplay between the features of the function approximator and the data distribution under the $\epsilon$-greedy exploratory policy with $\epsilon=0.05$:} (i) In early episodes, exploitation leads to an increase in the probability of sampling state-action pair ($s_1,a_1$) from the replay buffer since we estimate that $Q_w(s_1,a_1) > Q_w(s_1,a_2)$ (Fig. \ref{fig:four_states_mdp_exps_q_vals} (b)). This is clearly seen in Fig. \ref{fig:four_states_mdp_exps_2} (c), as $\mu(s_1,a_1)$ features a steep increase in probability during early training; (ii) the increase in probability $\mu(s_1,a_1)$, as well as a decrease in probability $\mu(s_2,a_1)$, drives an increase in weight $w_1$, as previously discussed; (iii) due to the fact that the target used in the estimation of $w_2$, associated with $Q_w(s_1,a_2)$, depends on $w_1$ (Eqs. \ref{eq:TD_weights_update}), the increase in $w_1$ will also drive an increase in $w_2$. However, the increase in weight $w_2$ is slower than the increase in weight $w_1$ because the pair $(s_1,a_2)$ is underrepresented in the replay buffer, i.e., the learning of $Q_w(s_1,a_2)$ occurs at a much slower pace than that of $Q_w(s_1,a_1)$. Nevertheless, happens that, when a certain threshold is surpassed, we wrongly estimate $Q_w(s_1,a_1) < Q_w(s_1,a_2)$, as can be seen in Fig. \ref{fig:four_states_mdp_exps_q_vals} (b) around episode 2 500. This leads to a drop in the obtained reward and number of correct actions, as seen in Fig. \ref{fig:four_states_mdp_exps}, due to the fact that action $a_2$ is wrongly taken at state $s_1$; (iv) since we now wrongly estimate $Q_w(s_1,a_1) < Q_w(s_1,a_2)$, exploitation leads to an increase in probabilities $\mu(s_1,a_2)$ and $\mu(s_2,a_1)$, driving weight $w_1$ down, until $\mu(s_1,a_1)$ is low enough such that we correctly estimate $Q_w(s_1,a_1) > Q_w(s_1,a_2)$ again. As can be seen in Fig. \ref{fig:four_states_mdp_exps_2} (c), after an initial increase, we observe a decrease in probability $\mu(s_1,a_1)$, as well as increase in probability $\mu(s_2,a_1)$, between episodes $2 500$ and $5 000$; The increase in probability $\mu(s_2,a_1)$ is driven by the fact that $Q_w(s_2,a_1) > Q_w(s_2,a_2)$ always verifies (Fig. \ref{fig:four_states_mdp_exps_q_vals} (b)); (v) the described interplay repeats again. However, the oscillation is now dampened by the fact that the replay buffer is already partially-filled with previous experience.

\subsubsection{Online Temporal Difference Version with Limited Replay Capacity}
\begin{figure*}[t]
    \centering
    \begin{subfigure}[t]{0.4\textwidth}
        \centering
        \includegraphics[width=0.99\textwidth]{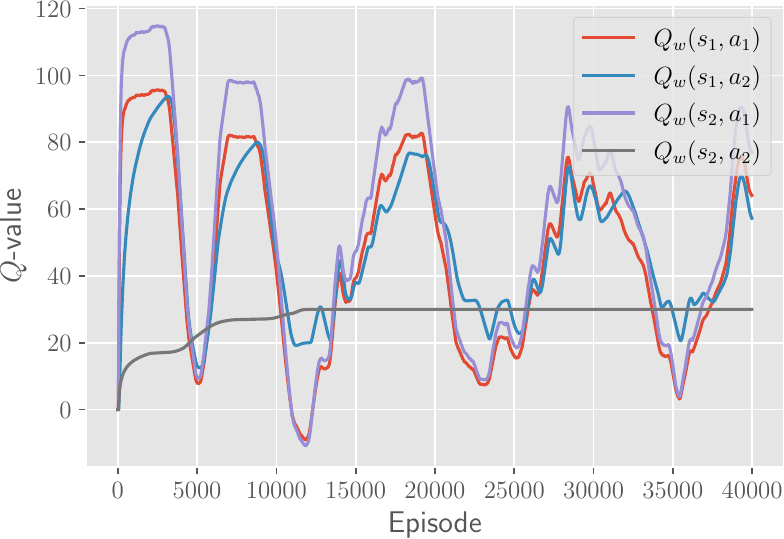}
        \caption{Replay buffer size = 10 000.}
    \end{subfigure}%
    ~ 
    \begin{subfigure}[t]{0.4\textwidth}
        \centering
        \includegraphics[width=0.99\textwidth]{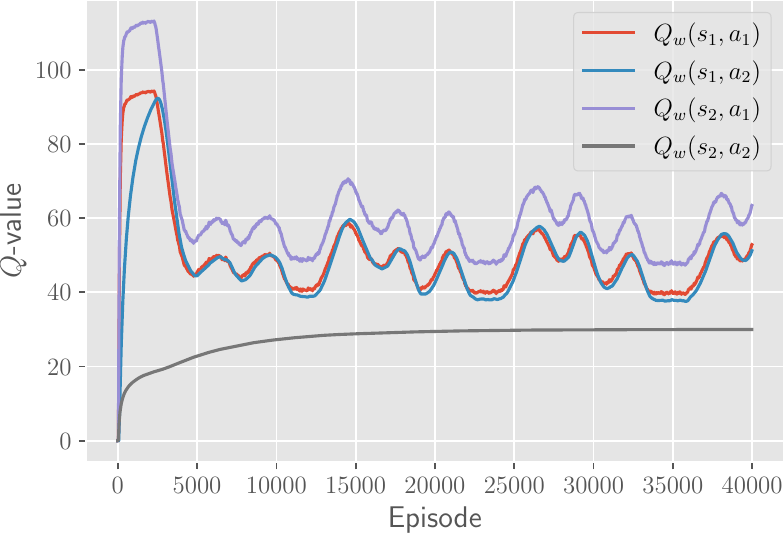}
        \caption{Replay buffer size = 50 000.}
    \end{subfigure}
    \begin{subfigure}[t]{0.4\textwidth}
        \centering
        \includegraphics[width=0.99\textwidth]{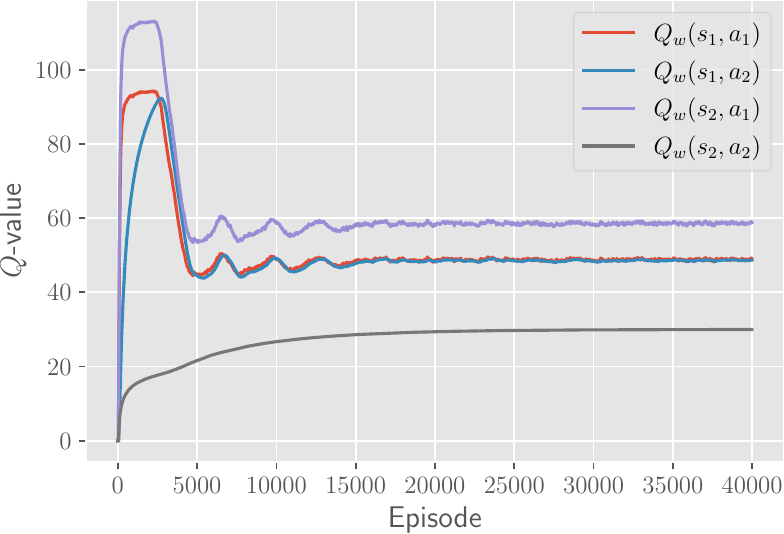}
        \caption{Replay buffer size = $\infty$.}
    \end{subfigure}
    \caption{Four-state MDP experiments for the $\epsilon$-greedy exploratory policy with $\epsilon=0.05$ under different replay buffer sizes. The plots display the estimated $Q$-values throughout episodes.}
    \label{fig:four_states_mdp_exps_replay_buffer}
\end{figure*}

\begin{figure*}[t]
    \centering
    \begin{subfigure}[t]{0.4\textwidth}
        \centering
        \includegraphics[width=0.99\textwidth]{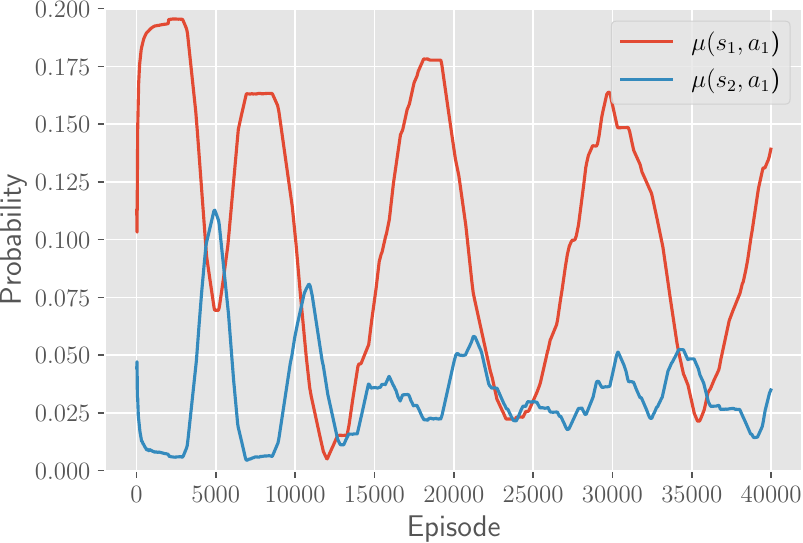}
        \caption{Replay buffer size = 10 000.}
    \end{subfigure}%
    ~ 
    \begin{subfigure}[t]{0.4\textwidth}
        \centering
        \includegraphics[width=0.99\textwidth]{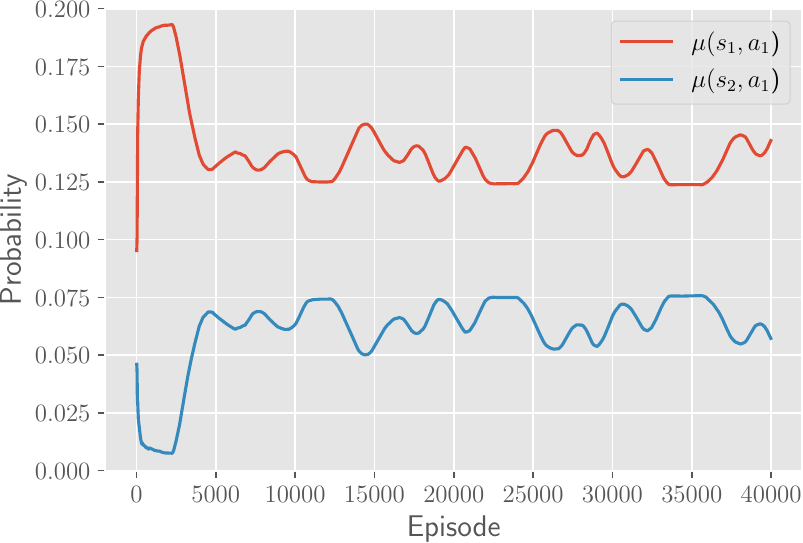}
        \caption{Replay buffer size = 50 000.}
    \end{subfigure}
    \begin{subfigure}[t]{0.4\textwidth}
        \centering
        \includegraphics[width=0.99\textwidth]{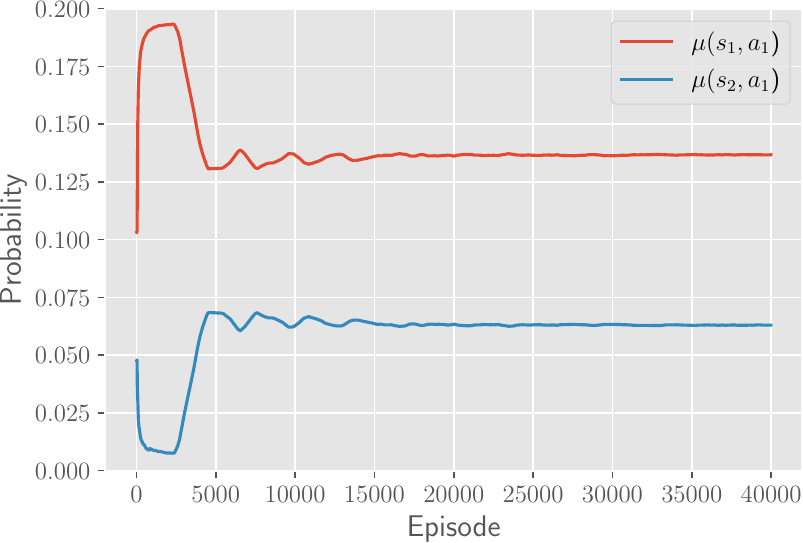}
        \caption{Replay buffer size = $\infty$.}
    \end{subfigure}
    \caption{Four-state MDP experiments for the $\epsilon$-greedy exploratory policy with $\epsilon=0.05$ under different replay buffer sizes. The plots display the data distribution probabilities $\mu(s_1,a_1)$ and $\mu(s_2,a_1)$, as induced by the contents of the replay buffer, throughout episodes.}
    \label{fig:four_states_mdp_exps_replay_buffer_2}
\end{figure*}

Lastly, we consider an experimental setting where the replay buffer has limited capacity. Figures \ref{fig:four_states_mdp_exps_replay_buffer} and \ref{fig:four_states_mdp_exps_replay_buffer_2} display the experimental results obtained with the $\epsilon$-greedy exploratory policy with $\epsilon=0.05$ by varying the size/capacity of the replay buffer. As can be seen in Fig. \ref{fig:four_states_mdp_exps_replay_buffer_2}, as the replay buffer size increases, the amplitude of the oscillations in the $\mu$ distribution gets smaller. Moreover, as the replay buffer size increases, the oscillations in the $Q$-values and $Q$-values errors are smaller, as seen in Fig. \ref{fig:four_states_mdp_exps_replay_buffer}. Given the previous discussion, the obtained results are expected. The undesirable interplay between the function approximator and the data distribution repeats as previously discussed for the infinitely-sized replay buffer. However, as the replay buffer gets smaller, the more the data distribution induced by the contents of the replay buffer is affected by changes to the current exploratory policy, in this case the $\epsilon$-greedy exploratory policy. Therefore, for smaller replay buffers, exploitation leads to more steep changes in $\mu(s_1,a_1)$ and $\mu(s_2,a_1)$ probabilities, as seen in Fig. \ref{fig:four_states_mdp_exps_replay_buffer_2} (a). Such changes in the $\mu$ distribution drive abrupt changes in weights $w_1$ and $w_2$, as well as estimated $Q$-values, as seen in Fig. \ref{fig:four_states_mdp_exps_replay_buffer} (a). Similarly to the infinitely-sized replay buffer, the agent keeps alternating between phases where it estimates that $Q_w(s_1,a_1) > Q_w(s_1,a_2)$ and phases where it estimates the opposite. However, the period at which phases switch is rather longer for smaller replay buffer sizes. Whereas for the infinitely-sized replay buffer the amplitude of the oscillations is dampened due to the fact that previously stored experience contributes to make the data distribution more stationary, this is not as easily achieved by smaller replay buffers. As our results suggest, the size of the replay buffer influences the stability of the data distribution, which can, in turn, affect the stability of the learning algorithm and quality of the resulting policies.

\subsubsection{Discussion}
In summary, this manuscript presented a set of experiments under a four-state MDP that show how the data distribution can greatly influence the performance of the resulting policies and the stability of the learning algorithm.
First, we showed that, under offline RL settings, the number of correct actions is directly dependent on the properties of the data distribution due to an undesirable correlation between features.
Second, under online RL settings, not only the quality of the retrieved policies depends on the data collection mechanism, but also an undesirable interplay between the data distribution and the function approximator can arise: exploitation can lead to abrupt changes in the data distribution, thus hindering learning. Additionally, we showed that the replay buffer size can also affect learning stability.

Above all, the results presented emphasize the key role played by the data distribution in the context of off-policy RL algorithms with function approximation. Despite the fact that we study a four-state MDP, we argue that the example here presented can generalize into more realistic settings. In the first place, it is possible to construct an MDP such that the learning dynamics under a function approximator with state-dependent features are equivalent to the previously discussed approximator with (state, action)-dependent features. %Figure \ref{fig:four_states_mdp_variation_appendix} displays an MDP whose learning dynamics under a function approximator with state-dependent features are equivalent to the learning dynamics of the four-state MDP under a function approximator with (state, action)-dependent features.
Second, we can abstract the example here presented by hypothesizing that states $s_1$ and $s_3$ can correspond to states along an optimal trajectory of a larger MDP, and state $s_2$ to a state outside the optimal trajectory. If the features of the states are wrongly correlated (as in the example, we have a correlation between the features of states $s_1$ and $s_2$), exploitation along the optimal trajectory can deteriorate the $Q$-values for states outside the optimal trajectory (as in the example, state $s_3$), which can give rise to algorithmic instabilities.

\section{Supplementary Materials for Section \ref{sec:experimental_results}}
\label{sec:supplementarysec5}
This section gives additional details with respect to Section \ref{sec:experimental_results}. The section is divided as follows:
\begin{itemize}
    \item \autoref{sec:supplementarysec5:expEnvs}: Experimental environments.
    \item \autoref{sec:supplementarysec5:expMethodology}: Experimental methodology.
    \item \autoref{sec:supplementarysec5:hyperparameters}: Implementation details and algorithms' hyperparameters.
    \item \autoref{sec:supplementarysec5:completeExpResults}: Complete experimental results.
\end{itemize}

\subsection{Experimental Environments}
\label{sec:supplementarysec5:expEnvs}
Below, we detail the experimental environments used in this work.
\subsubsection{Grid 1 and Grid 2 Environments}
\begin{figure}[h]
\begin{center}
    \begin{tabular}{cc}
    \includegraphics[width=0.24\textwidth]{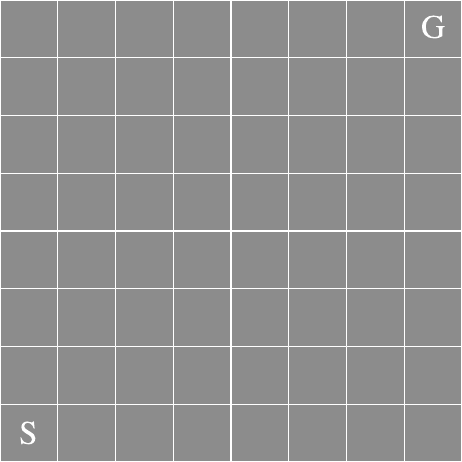} \hspace{1cm} &
    \includegraphics[width=0.24\textwidth]{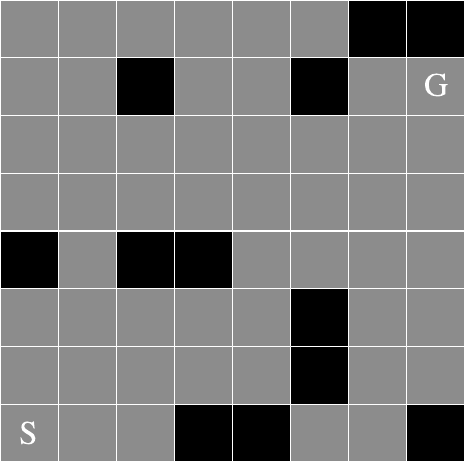} \\
    (a) Grid 1 environment. \hspace{1cm} & (b) Grid 2 environment.  \\
    \end{tabular}
\caption{Illustration of the grid environments.}
\label{fig:grid_envs}
\end{center}
\end{figure}

The grid 1 environment comprises a tabular $8 \times 8$ grid, as seen in Fig. \ref{fig:grid_envs} (a). The agent starts in the lower left corner (``S'' square) and aims to reach the upper right corner (``G'' square) as fast as possible. The grid 2 environment, as seen in Fig. \ref{fig:grid_envs} (b), comprises a tabular $8 \times 8$ grid, with walls in the middle of the environment. The agent starts in the lower left square (``S'' square) and aims to reach the upper right square (``G'' square) as fast as possible. 

Common to both environments, the episodes have a fixed length of 50 timesteps. The action set is \{up, down, right, left, stay\}. No diagonal movements are allowed. All actions lead to deterministic state transitions except for against-wall actions (the environments are virtually delimited by four walls); in such cases the agent has a 0.01 probability of moving to an adjacent square and with the remainder probability stays at the same square. The reward at each timestep is one if the agent is at the goal state and zero otherwise. The underlying MDPs comprise $(8\times 8) \times 5 = 320$ state-action pairs.

Regarding the state features, each state is mapped to a $8$-dimensional vector. Each entry of the vector is drawn, independently, from an uniform distribution in $[-1,1]$. The state features are pre-computed during initialization and kept constant throughout training. We set the discount factor $\gamma$ to 0.9.

\subsubsection{Multi-path Environment}
\begin{figure}[h]
\begin{center}
    \includegraphics[width=0.6\textwidth]{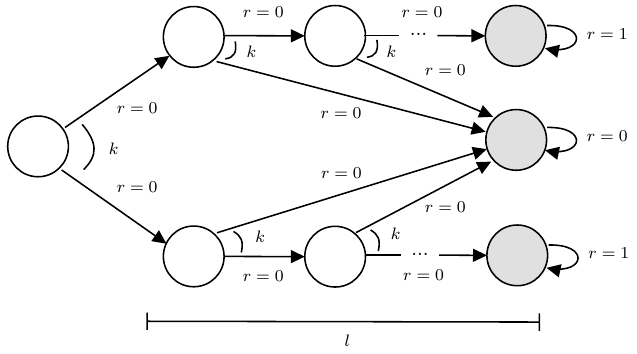}
    \caption{Illustration of the multi-path ($k,l$) environment.}
    \label{fig:multi_paths_env}
\end{center}
\end{figure}

The multi-path environment ($k,l)$, displayed in Fig. \ref{fig:multi_paths_env}, comprises $k$ parallel paths of length $l$.  The agent starts in the left-most state and aims to reach one of the $k$ absorbing states with positive reward on the right, while avoiding to fall into the absorbing state with zero reward. In other words, the agent aims at following one of the $k$ paths all the way until the positive-rewarding state without falling into the absorbing zero reward state.

The action set is $\{1,2,3,...,k\}$. All actions lead to deterministic state transitions except for the action taken in the initial state; at the initial state, the agent's action succeeds with probability $1-p$ and with probability $p$, the agent will randomly enter the first state of one of the $k$ paths. By default we set $p=0.01$ The reward is one if the agent is at one of the $k$ final absorbing states of the path and zero everywhere else. The episode length is 10 timesteps. By default, we set $k=5$ and $l=5$. Therefore, the underlying MDP comprises $(k\times l+2)\times k=135$ state-action pairs.

Regarding the state features, each state is mapped to a $4$-dimensional vector. Each entry of the vector is drawn from an uniform distribution in $[-1,1]$. The state features are pre-computed during initialization and kept constant throughout training. We set the discount factor $\gamma$ to 0.9.

\subsubsection{Pendulum, Cartpole and Mountaincar Environments}
The pendulum, cartpole and mountaincar environments are based on the OpenAI gym environments implementations. For all environments, we set the discount factor $\gamma$ to 0.99. For the pendulum environment, the action space is discretized into 5 equally-spaced discrete actions.

For all environment, and only with the purpose of calculating different properties of the datasets (such as coverage, or entropy), we discretized the state-space. We used an histogram-like discretization with equally spaced bins. For both the pendulum and mountaincar environments we discretized each dimension of the state-space using 50 bins; for the cartpole environment we used 20 bins per state-space dimension.

\subsection{Experimental Methodology}
\label{sec:supplementarysec5:expMethodology}
In order to reliably compare the different algorithms used in this work, we followed a rigorous experimental evaluation methodology. Under each experimental configuration, we perform 4 training runs. Each training run is punctually evaluated during training, as well as at the end of training, by running 5 evaluation rollouts. Each rollout comprises a full-length episode. We then average the resulting $(4 \times 5)$ performance indicators per train run, thus obtaining 4 performance samples. We focus our attention on two main types of performance indicators: (i) the reward obtained under evaluation rollouts; and (ii) the $Q$-values errors, calculated using a value-iteration or $Q$-learning solution for the same MDP;

% We report the pointwise estimation of the mean values for the different performance indicators. Additionally, we provide 95\% bootstrapped confidence intervals, calculated using 25 000 resamples. We visually represent the confidence interval in plots with shaded regions or vertical bars. %We additionally report the optimality gap \citep{agarwal_2021}.

% With respect to non-pointwise metrics, we display distribution plots that can be used to get a full picture of the distribution underlying the pointwise metrics.

\subsection{Implementation Details and Hyperparameters}
\label{sec:supplementarysec5:hyperparameters}
We developed our software in a Python environment, using the following additional open-source frameworks: ACME \citep{hoffman2020acme}, and Tensorflow \citep{tensorflow}. Table \ref{tab:supplementarysec5:dqn_hyperparameters} presents the default hyperparameters for the offline DQN and CQL algorithms used in the experiments presented in Section \ref{sec:experimental_results}.

\begin{table}[h]
    \caption{Algorithms' hyperparameters used in Sec. \ref{sec:experimental_results}.}
    \label{tab:supplementarysec5:dqn_hyperparameters}
    \begin{center}
        \begin{tabular}{lccc}
        \hline 
                                       & \textbf{Grid 1, 2 and}  & \textbf{Pendulum and}  & \textbf{Cartpole} \\
        \textbf{Hyperparameter}        & \textbf{multi-path envs.}  & \textbf{Mountaincar envs.}  & \textbf{env.} \\ \hline
        Optimizer                       & Adam        & Adam           & Adam \\ 
        Learning rate                   & 1e-3         & 1e-3   & 1e-3 \\ 
        Discount factor ($\gamma$)      & 0.9        & 0.99         & 0.99 \\
        Dataset size                    & 50 000     & 200 000     & 1 000 000 \\
        Batch size                      & 100        & 100          & 100   \\
        Target update period            & 1 000      & 1 000   & 1 000 \\
        Num. learning steps             & 100 000    & 200 000  & 500 000 \\
        Network layers                  & [20,40,20] & [64,128,64] & [64,128,64] \\
        Alpha (CQL param.)              & 1.0        & 1.0    & 1.0 \\
        \end{tabular}
    \end{center}
\end{table}

\subsection{Complete Experimental Results}
\label{sec:supplementarysec5:completeExpResults}
The complete experimental results can be found in the following interactive dashboard: \url{https://rldatadistribution.pythonanywhere.com/}.

\end{appendices}

\end{document}